\pdfoutput=1

\documentclass[11pt]{article}

\usepackage[preprint]{acl}

\usepackage{times}
\usepackage{latexsym}

\usepackage[T1]{fontenc}

\usepackage[utf8]{inputenc}

\usepackage{microtype}

\usepackage{inconsolata}

\usepackage{graphicx}
\usepackage{amsmath}
\usepackage{subfigure}
\usepackage{float}
\usepackage{multirow}
\usepackage{makecell}
\usepackage{booktabs}
\usepackage{amsfonts}
\usepackage{colortbl}
\usepackage{enumerate}
\title{Adapt Once, Thrive with Updates: Transferable Parameter-Efficient Fine-Tuning on Evolving Base Models}

\author{Naibin Gu\textsuperscript{\rm 1,2},\ Peng Fu\textsuperscript{\rm 1,2}\thanks{\ \  Corresponding author: Peng Fu.},\ Xiyu Liu\textsuperscript{\rm 1,2},\ Ke Ma\textsuperscript{\rm 3},\ Zheng Lin\textsuperscript{\rm 1,2},\ Weiping Wang\textsuperscript{\rm 1} \\ 
\textsuperscript{\rm 1}Institute of Information Engineering, Chinese Academy of Sciences, Beijing, China \\
\textsuperscript{\rm 2}School of Cyber Security, University of Chinese Academy of Sciences, Beijing, China \\
\textsuperscript{\rm 3}School of Electronic, Electrical and Communication Engineering, UCAS, Beijing, China \\
  \texttt{\textrm{\{}gunaibin,fupeng\textrm{\}}@iie.ac.cn} \\
  }

\begin{document}
\maketitle
\begin{abstract}
Parameter-efficient fine-tuning (PEFT) has become a common method for fine-tuning large language models, where a base model can serve multiple users through PEFT module switching. To enhance user experience, base models require periodic updates. However, once updated, PEFT modules fine-tuned on previous versions often suffer substantial performance degradation on newer versions. Re-tuning these numerous modules to restore performance would incur significant computational costs. Through a comprehensive analysis of the changes that occur during base model updates, we uncover an interesting phenomenon: continual training primarily affects task-specific knowledge stored in Feed-Forward Networks (FFN), while having less impact on the task-specific pattern in the Attention mechanism. Based on these findings, we introduce Trans-PEFT, a novel approach that enhances the PEFT module by focusing on the task-specific pattern while reducing its dependence on certain knowledge in the base model. Further theoretical analysis supports our approach. Extensive experiments across 7 base models and 12 datasets demonstrate that Trans-PEFT trained modules can maintain performance on updated base models without re-tuning, significantly reducing maintenance overhead in real-world applications\footnote{The code is available at \url{https://github.com/gccnlp/Trans-PEFT}.}.
\end{abstract}

\section{Introduction}
Large language models have shown impressive performance across various domains \citep{DBLP:journals/corr/abs-2307-09288, OpenAI2023GPT4TR, yang2024qwen2technicalreport,chen2024mixturehiddendimensionstransformer,cai2024internlm2technicalreport}. While fine-tuning is a common approach to adapt these models for specific applications, it becomes increasingly challenging in terms of computational resources and storage requirements as model sizes grow. Parameter-efficient fine-tuning (PEFT) techniques \citep{DBLP:conf/icml/HoulsbyGJMLGAG19, DBLP:conf/iclr/HuSWALWWC22, gu-etal-2024-light,pmlr-v235-liu24bn} address this challenge by updating only a small subset of parameters while maintaining comparable performance. This enables efficient model customization where a single base model can serve numerous users through dynamic switching between PEFT modules.

In order to enhance user experience, base models require periodic updates to update their knowledge and improve their capabilities. These updates, which do not alter the architecture, are typically achieved through continual pre-training, as exemplified by the updating from Qwen2 \citep{yang2024qwen2technicalreport} to Qwen2.5 \citep{qwen2025qwen25technicalreport}. However, a critical challenge emerges: when the base model is updated, the PEFT modules fine-tuned on the previous version cannot be applied to the new version, as such direct transfer would result in a significant drop in performance~\cite{qin-etal-2023-recyclable}.  For large-scale deployments with numerous PEFT modules, re-tuning not only incurs substantial computational overhead but also raises significant privacy concerns due to the necessity of maintaining long-term storage of users' data.

To tackle the challenge of performance degradation after transferring, we conduct an analysis of the changes that occur during base model updates and uncover several interesting observations: (1) Attention activation distributions remain similar in corresponding layers, suggesting that fine-tuned base models maintain similar task-specific 
patterns across versions. (2) FFN activation distributions differ between versions in corresponding layers, suggesting changes in task-specific knowledge storage within layers. (3) The influence of different FFN sub-layers on activation magnitudes changes, indicating shifts in both intra-layer and cross-layer knowledge storage. These changes in knowledge storage lead to a decline in task performance after transferring, highlighting the need to reduce dependence on certain knowledge.

Building upon these findings, we introduce Trans-PEFT, a novel approach to enhance PEFT module transferability. During fine-tuning, Trans-PEFT employs two key strategies: Intra-layer Knowledge Masking and Cross-layer Knowledge Dropping. These strategies systematically reduce the dependence of PEFT modules on certain knowledge, thereby encouraging PEFT to capture the invariant task-specific patterns within the attention mechanisms. Furthermore, we provide a theoretical guarantee for the task performance after transferring by bounding the loss discrepancy between base model versions. In practical applications, Trans-PEFT can seamlessly integrate with existing PEFT methods without requiring architectural modifications or introducing additional computational costs. This practical design allows PEFT modules trained on old base models to be efficiently transferred to their new versions.

We validate our approach using 7 different base models from 3 different sources, across 12 datasets covering math reasoning, code generation, and commonsense reasoning. Experimental results demonstrate that Trans-PEFT maintains the performance across base model versions without re-tuning, achieving up to 30\% performance gains compared to Direct Transfer. Furthermore, Trans-PEFT exhibits the capability not only to maintain performance but also to potentially leverage the improvements introduced by base model updates.

In summary, our contributions are as follows:
\begin{itemize}
\item We reveal the impact of base model updates: continual training significantly modifies task-specific knowledge stored in the FFN sub-layer, while it has a minor impact on the task-specific patterns of the attention mechanism.
\item We introduce Trans-PEFT, a novel method enabling the transfer of PEFT modules from old base model versions to their new versions without re-tuning. We provide a theoretical analysis to establish the foundations of its effectiveness.
\item Extensive experiments on 7 base models across 12 datasets show that Trans-PEFT trained modules can be used directly without re-tuning while maintaining performance.
\end{itemize}

\section{Preliminary}
\begin{figure}[t]
        \centering
	\subfigure[InternLM2-7B]{
	\includegraphics[width=0.48\linewidth]{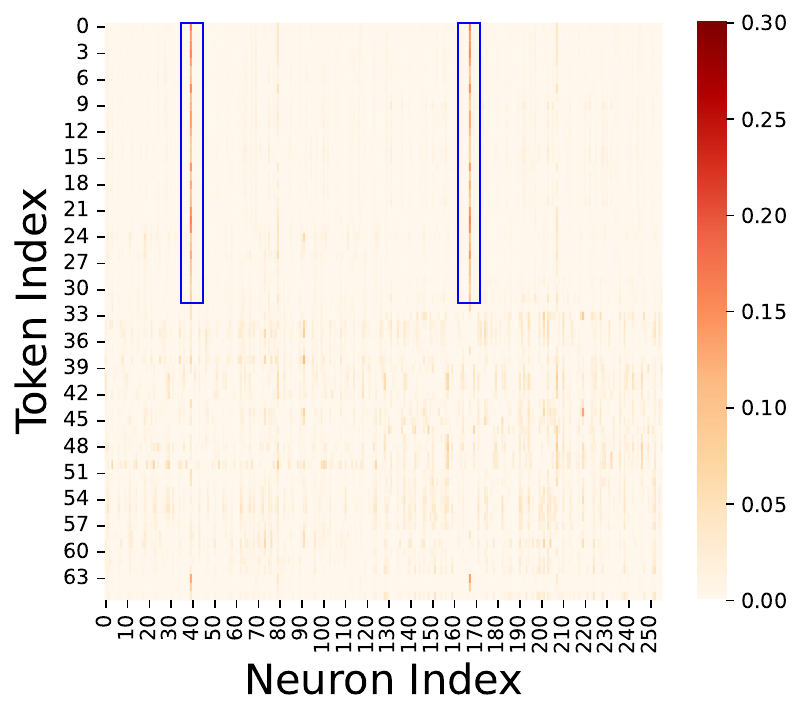}}
        \subfigure[InternLM2.5-7B]{
	\includegraphics[width=0.48\linewidth]{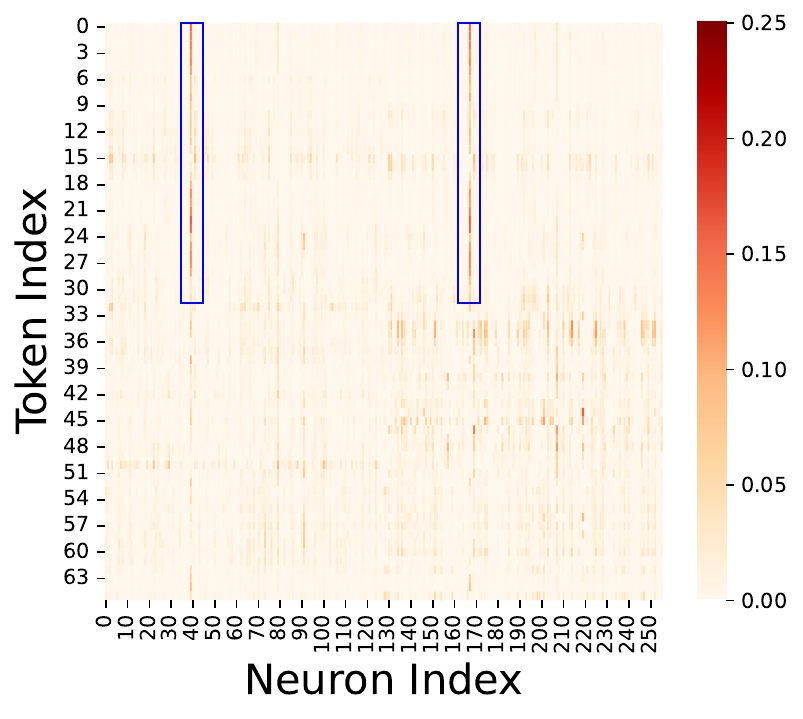}} 
        \newline
         \subfigure[Qwen2-7B]{
	\includegraphics[width=0.48\linewidth]{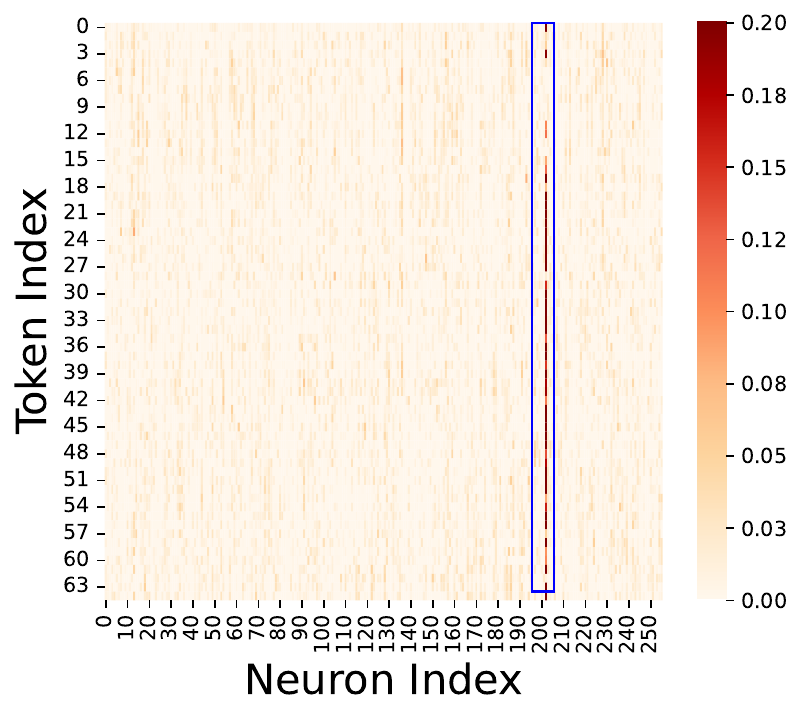}}
        \subfigure[Qwen2.5-7B]{
	\includegraphics[width=0.48\linewidth]{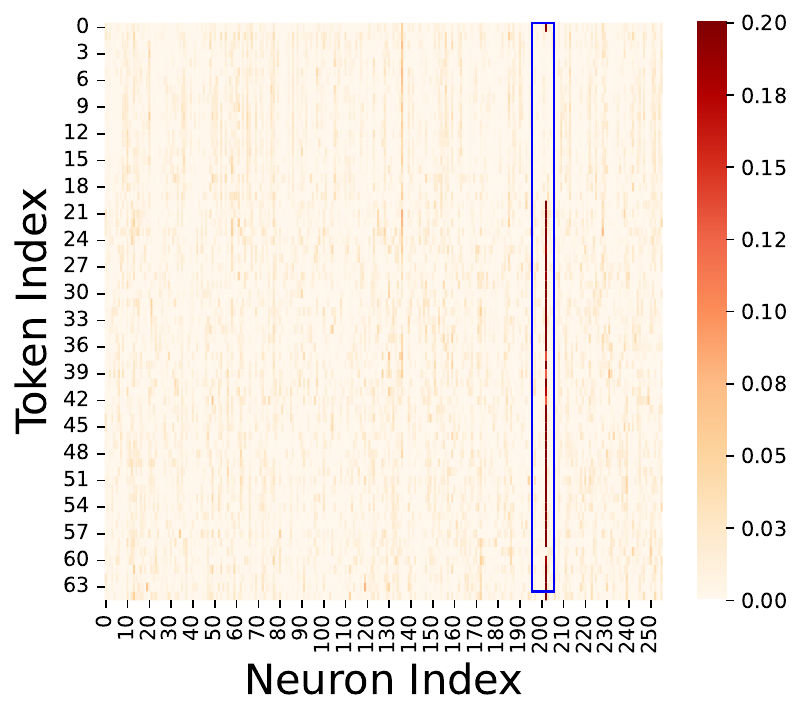}} 
\caption{Comparison of activation distributions within attention sub-layers across different fine-tuned base model versions. The visualization shows the same dimensions from the 18th layer of four base models, with similar regularities observed across other layers.}
\label{pilot-attn}
\end{figure}
\begin{figure}[t]
        \centering
	\subfigure[InternLM2-7B]{
	\includegraphics[width=0.48\linewidth]{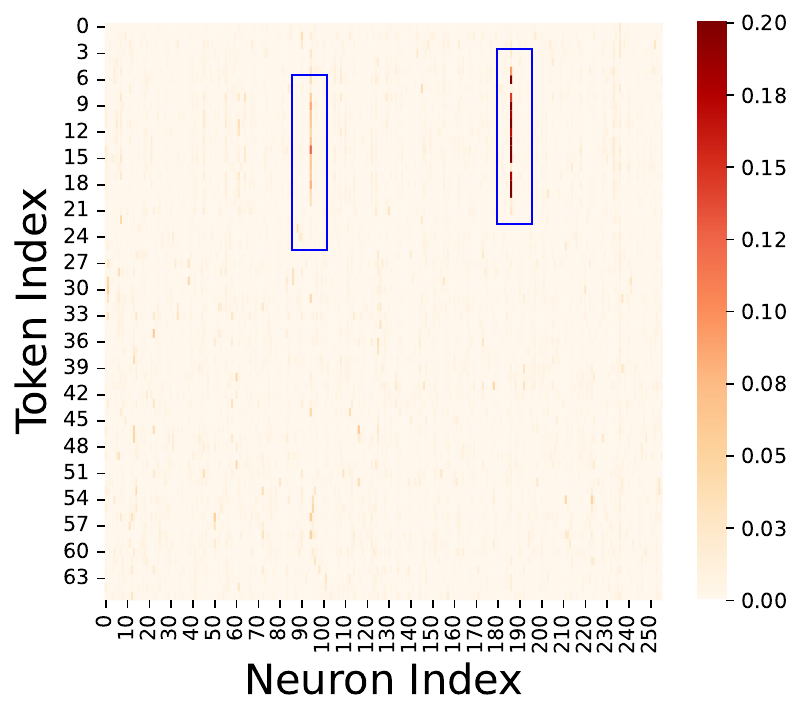}} 
        \subfigure[InternLM2.5-7B]{
	\includegraphics[width=0.48\linewidth]{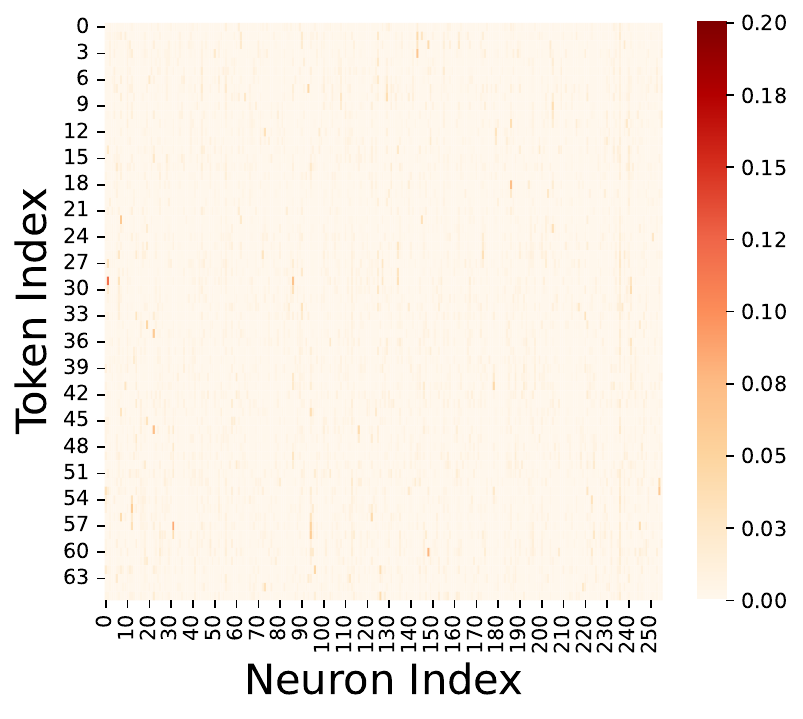}}
        \newline
        \subfigure[Qwen2-7B]{
	\includegraphics[width=0.48\linewidth]{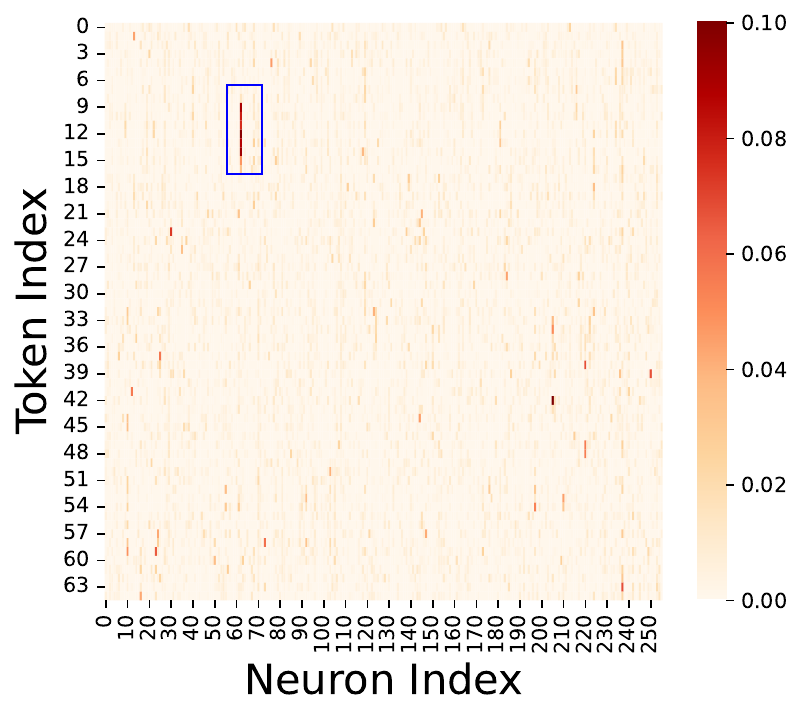}}
        \subfigure[Qwen2.5-7B]{
	\includegraphics[width=0.48\linewidth]{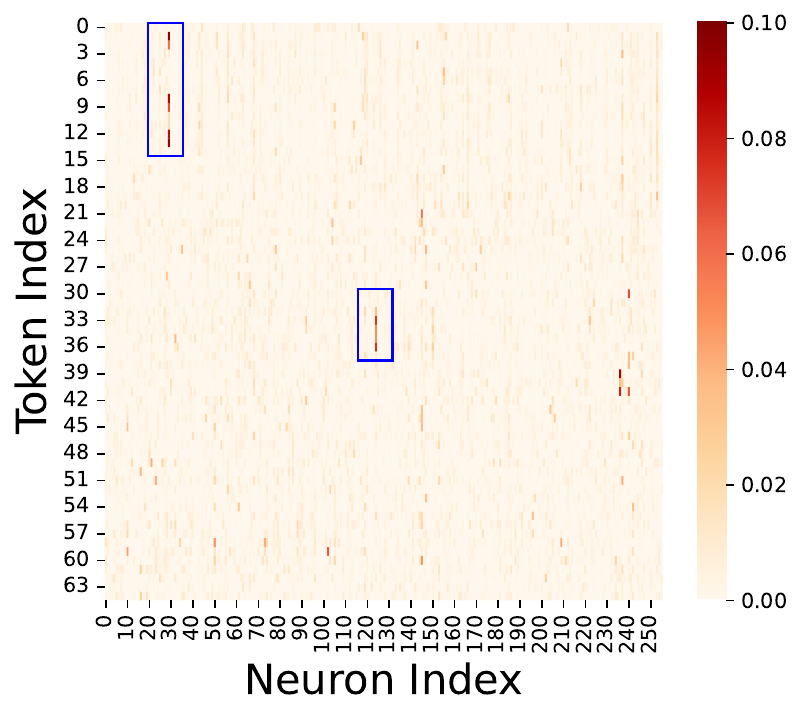}} 
\caption{Comparison of activation distributions within FFN sub-layers across different fine-tuned base model versions. The visualization shows the same dimensions from the 18th layer of four base models, with similar regularities observed across other layers.}
\label{pilot-ffn}
\end{figure}
In our Trans-PEFT approach, we select two widely used PEFT techniques, LoRA \citep{DBLP:conf/iclr/HuSWALWWC22} and Adapter \citep{DBLP:conf/icml/HoulsbyGJMLGAG19}, to validate the effectiveness of transferability.

\noindent\textbf{LoRA.} For a linear weight $\mathbf{W}\in{\mathbb{R}^{d\times k}}$ in the base model, LoRA introduces the trainable module in parallel with the linear weight. Each module consists of a down-projection layer $\mathbf{W}_{\text{down}}\in{\mathbb{R}^{d\times r}}$ and an up-projection layer $\mathbf{W}_{\text{up}}\in{\mathbb{R}^{r\times k}}$, where $d$ is the hidden size of the base model and $r\ll d$. After incorporating LoRA, the forward pass is modified as follows:
\begin{equation}
    \mathbf{h}\leftarrow \mathbf{X}\mathbf{W}+\mathbf{X}\mathbf{W}_{\text{down}}\mathbf{W}_{\text{up}},
\end{equation}
where $\mathbf{X}\in{\mathbb{R}^{d}}$ is the input of the $\mathbf{W}$.

\noindent\textbf{Adapter.} For each layer in the base model, the Adapter method adds two trainable modules sequentially after the attention sub-layer and the FFN sub-layer. After adding adapters, the forward pass is modified as follows:
\begin{equation}
    \mathbf{h}\leftarrow \mathbf{h}+f(\mathbf{h}\mathbf{W}_{\text{down}})\mathbf{W}_{\text{up}},
\end{equation}
where $h\in{\mathbb{R}^{d}}$ is the output of the sub-layer, $f$ is a non-linear activation function.

\section{Analysis of Updating Base Models}
\begin{figure}[t]
        \centering
	\subfigure[InternLM2-7B vs InternLM2.5-7B]{
	\includegraphics[width=0.98\linewidth]{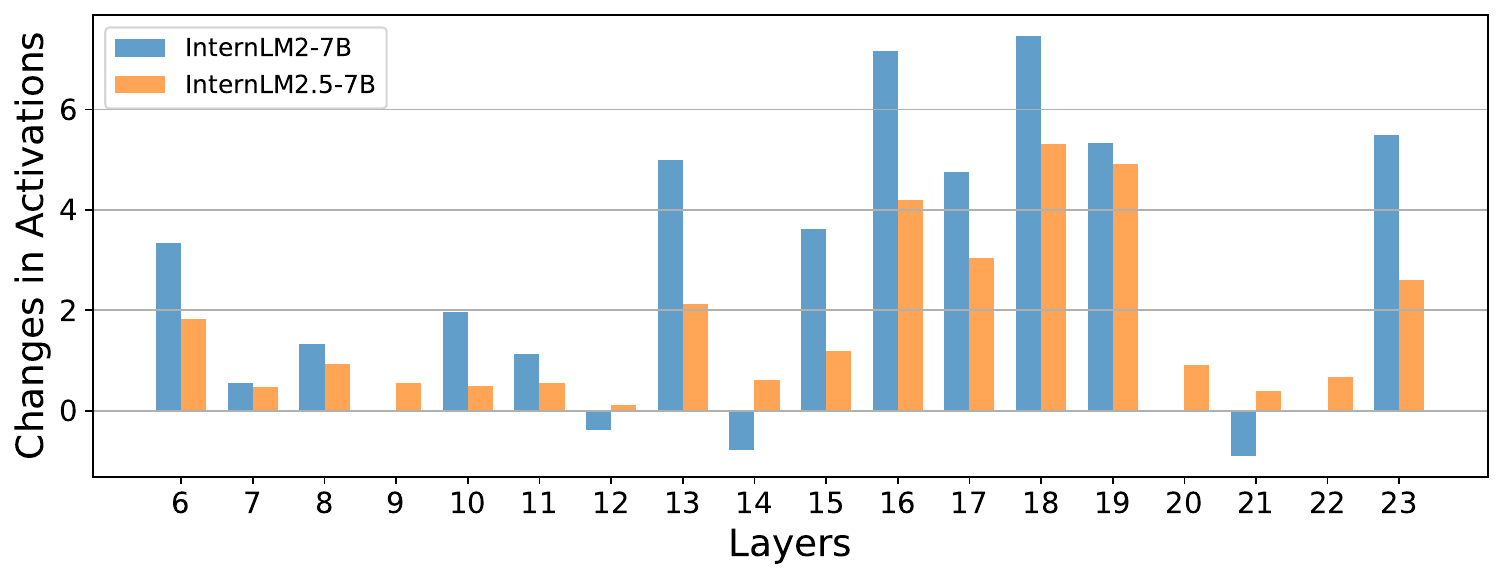}} 
        \newline
        \subfigure[Qwen2-7B vs Qwen2.5-7B]{
	\includegraphics[width=0.98\linewidth]{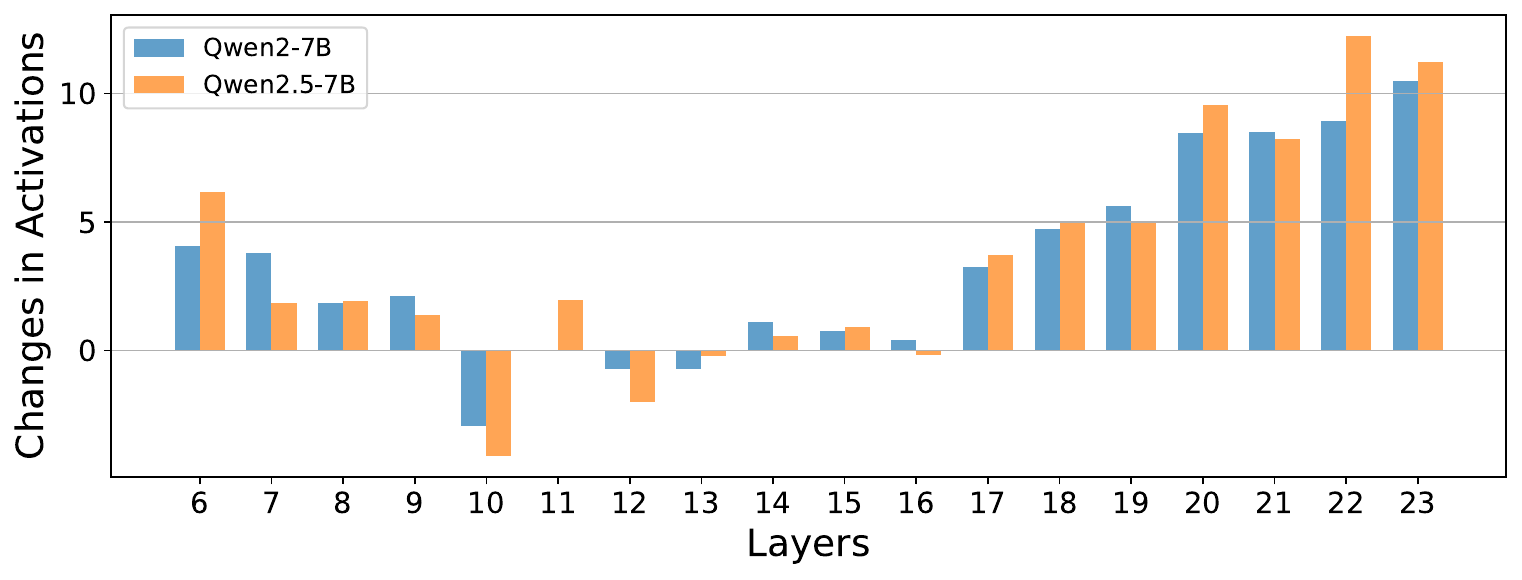}} 
\caption{Comparison of each FFN sub-layer's influence on activations across different fine-tuned base model versions. The influence is measured by the difference in the magnitude of activations between adjacent layers.}
\label{pilot-layer}
\end{figure}
\label{sec-pilot}
Modern base models typically employ the Transformer~\cite{DBLP:journals/corr/VaswaniSPUJGKP17} architecture, which consists of multiple transformer layers, with each containing an attention sub-layer and an FFN sub-layer. To investigate the impact of updating, we fine-tune LoRA on Qwen2-7B and its newer version Qwen2.5-7B, InternLM2-7B \citep{cai2024internlm2technicalreport} and its newer version InternLM2.5-7B with the MetaMathQA dataset~\citep{yu2024metamath} and analyze the activations in sub-layers of these models. See Appendix~\ref{sec-imp} for more implementation details.
\begin{figure*}[t]
    \centering
    \includegraphics[width=2\columnwidth,trim=52 150 52 150,clip]{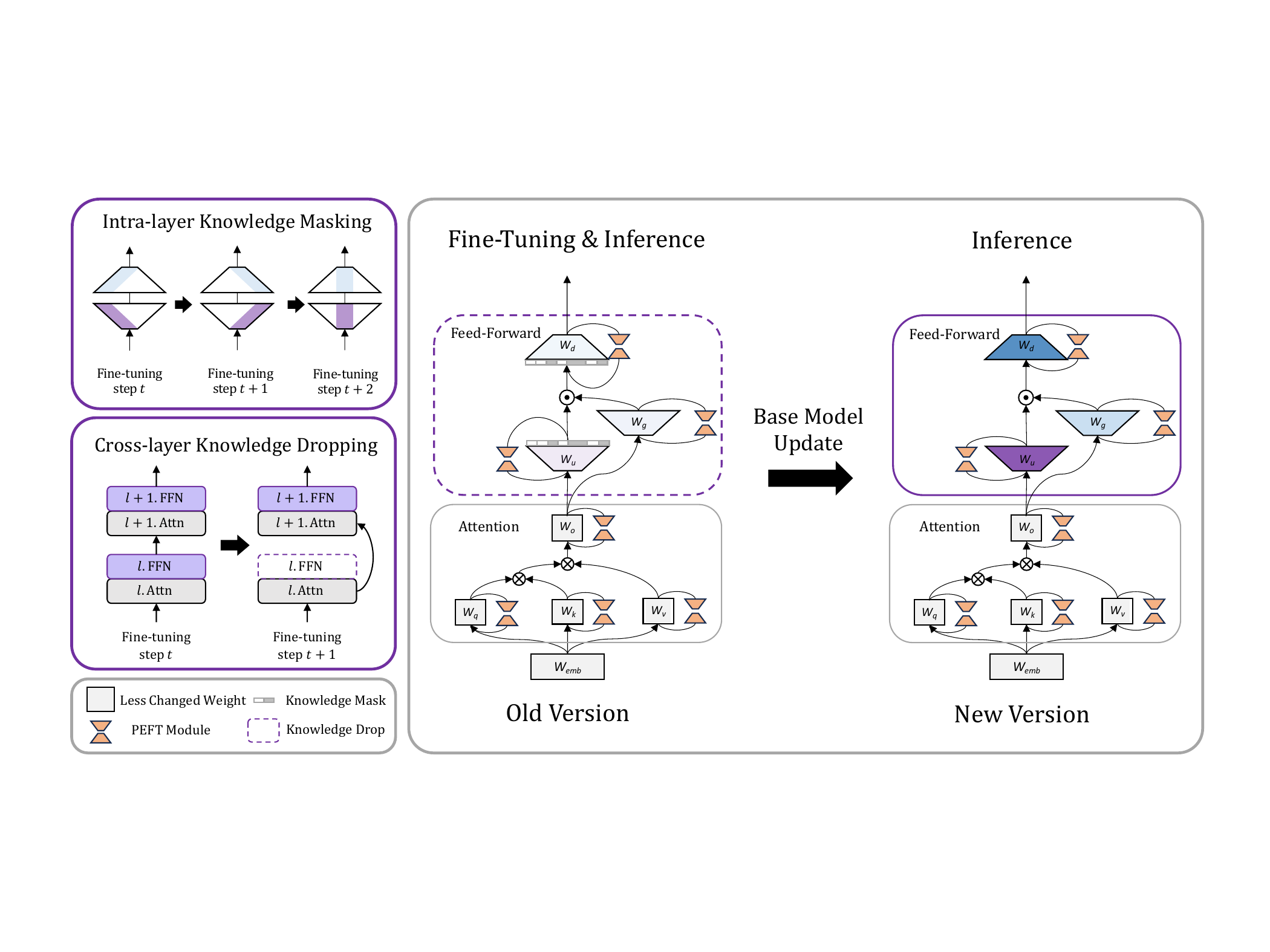}
    \caption{Illustration of Trans-PEFT. The Trans-PEFT approach mainly includes two strategies: intra-layer knowledge masking and cross-layer knowledge dropping. In each forward pass, it randomly chooses to drop the output of an entire FFN layer or to mask certain intermediate dimensions.}
    \label{method}
\end{figure*}
\subsection{Changes Occurring within the Layer}
We compare the activation distributions of two fine-tuned model versions for the attention and FFN sub-layers in each layer, as shown in Figures~\ref{pilot-attn} and~\ref{pilot-ffn}.
Blue rectangles highlight regions with relatively high activations. We find that after fine-tuning, the activation distributions of the attention sub-layer in both versions are quite similar. However, we observe notable differences in the intermediate activation distributions of the FFN sub-layer. Previous interpretability studies suggest that the primary role of the attention sub-layer is to extract dependencies between tokens \citep{hao2021selfattentionattributioninterpretinginformation}, while the FFN sub-layer primarily functions as key-value memories to store knowledge \citep{geva-etal-2021-transformer}. These results indicate that \textbf{after fine-tuning, the attention mechanism in both versions undergoes minimal changes regarding task-specific patterns for capturing token relationships}. Typically, updates to the base model without changing its architecture aim to change model knowledge and enhance capabilities through continual training. \textbf{The differences in FFN activations confirm that the two versions differ in the task-specific knowledge regions.} 
\subsection{Changes Occurring across Different Layers}
\label{obs-layer}
Furthermore, we investigate the knowledge storage changes across different FFN sub-layers. As illustrated in Figure~\ref{pilot-layer}, we analyze the influence of each FFN sub-layer on the activation magnitudes for both the old and new versions, which reflects modifications in task-specific knowledge storage. It is observed that many of the layers even exhibit opposing effects between the two versions, \textbf{suggesting knowledge storage also changes between different FFN sub-layers}. In summary, the unsuccessful direct transfer of PEFT modules to the updated base model can be attributed to two reasons: changes in both intra-layer and cross-layer knowledge storage of the base model.

\section{Methodology}
\subsection{Trans-PEFT}
Although we have discovered that transfer performance is influenced by redistributed knowledge storage in FFN sub-layers, we are unable to precisely track how the base model modifies its storage for different tasks.
This makes it impossible to determine the exact modifications needed for PEFT modules on their fine-tuning task. 

To address this challenge, we propose the Trans-PEFT approach, which improves the transferability of PEFT modules by modifying the fine-tuning process on the old version of the model. 
As illustrated in Figure~\ref{method}, our approach primarily incorporates two key strategies: Intra-layer knowledge masking and Cross-layer knowledge dropping.

\noindent\textbf{Intra-layer Knowledge Masking.}
Due to the uncertainty in knowledge storage changes caused by updates, we are motivated to introduce randomness to reduce dependency on certain knowledge. Specifically, given a FFN sub-layer, it primarily consists of an up-projection matrix $\mathbf{W}_{\text{fc1}}\in{\mathbb{R}^{d\times d_{ff}}}$, a down-projection matrix $\mathbf{W}_{\text{fc2}}\in{\mathbb{R}^{d_{ff}\times d}}$, and an activation function $\sigma$, where $d_{ff}$ is the intermediate size\footnote{The detailed description of the gate\_proj in LLaMA-style models are omitted for simplicity.}. When PEFT modules are added, taking LoRA as an example, the output for a given input $\mathbf{X}$ is:
\begin{equation}
\begin{aligned}
    \text{FFN}(\mathbf{X}) = & \, \sigma(\mathbf{X}(\mathbf{W}_{\text{fc1}} + {\Delta \mathbf{W}_{\text{fc1}}})) \\
    & \cdot (\mathbf{W}_{\text{fc2}} + {\Delta \mathbf{W}_{\text{fc2}}}),
\end{aligned}
\end{equation}
where ${\Delta \mathbf{W}}$ represents the added LoRA module on the corresponding weights.

To achieve this goal, we introduce a non-trainable binary mask vector $m$ for each FFN sub-layer during fine-tuning on the old base model version. This mask vector is applied element-wise to the output of the activation function, reducing its dependency on specific dimensions. The values of this mask vector are determined based on a rate $p_{i}$ as follows:
\begin{equation}
m \sim \text{Bernoulli}(1-p_{i}),
\end{equation}
\begin{equation}
\begin{aligned}
    \text{FFN}(\mathbf{X}) = & \, \sigma(\mathbf{X}(\mathbf{W}_{\text{fc1}} + {\Delta \mathbf{W}_{\text{fc1}}})) \odot m \\
    & \cdot (\mathbf{W}_{\text{fc2}} + {\Delta \mathbf{W}_{\text{fc2}}}),
\end{aligned}
\end{equation}
where $m$ varies with each forward pass, randomly masking the outputs of the intermediate dimensions. 

\noindent\textbf{Cross-layer Knowledge Dropping.} In addition to the intra-layer perspective strategy, based on our findings in Section~\ref{obs-layer}, we introduce randomness from a cross-layer perspective to alleviate the PEFT module’s dependency on specific layers' knowledge. Specifically, we introduce a binary element $z$ before each FFN layer to decide whether to drop the output of that layer. For each layer, the value of $z$ is determined based on the rate $p_{c}$ as follows:
\begin{equation}
z \sim \text{Bernoulli}(1-p_{c}),
\end{equation}
\begin{equation}
\begin{aligned}
      \widetilde{\text{FFN}}(\mathbf{X}) = z \cdot  \text{FFN}(\mathbf{X}),
\end{aligned}
\end{equation}
where $ \widetilde{\text{FFN}}(\mathbf{X})$ represents the final output of the FFN layer. This strategy introduces variability in which layers contribute to the final output, reducing PEFT modules’ dependency on any specific layer.

By integrating these two strategies, the intra-layer masking introduces fine-grained randomness, ensuring that the PEFT module does not overly rely on certain knowledge within the FFN layer. Meanwhile, the cross-layer knowledge dropping prevents the PEFT module from becoming overly dependent on the knowledge of certain layers, thereby encouraging PEFT modules to capture persistent task-specific patterns within the attention mechanisms. These strategies together enable PEFT modules to maintain effectiveness across different versions of the base model.
\subsection{Theoretical Analysis}
To provide theoretical insights into the effectiveness of our approach, we analyze how Trans-PEFT influences the transferability of PEFT modules across different model versions.

Let \( \mathcal{M}_0 \) and \( \mathcal{M}_1 \) denote two versions of a base transformer model, where \( \mathcal{M}_1 \) is updated from \( \mathcal{M}_0 \). We define \( \mathbf{W}_{\text{ffn}} = \{ \mathbf{W}_{\text{fc1}}, \mathbf{W}_{\text{fc2}} \} \) as the weights in FFN sub-layers, \( \theta_{\text{ffn}} = \{ \Delta\mathbf{W}_{\text{fc1}}, \Delta\mathbf{W}_{\text{fc2}} \} \) as the PEFT parameters for FFN sub-layers, \( \mathbf{W}_{\text{att}} \) as the attention sub-layer weights, and \( \Delta\mathbf{W}_{\text{att}} \) as the corresponding PEFT parameters. For an input \( \mathbf{X} \in \mathbb{R}^d \), the output of a transformer layer can be expressed as:
\begin{equation}
y = \text{Attn}(\mathbf{X}) + \text{FFN}(\text{Attn}(\mathbf{X})),
\end{equation}
where \( \text{Attn}(\cdot) \) and \( \text{FFN}(\cdot) \) denote the attention and FFN operations, respectively. When integrating PEFT modules (e.g., LoRA), the output becomes: 
\begin{equation}
y = \text{Attn}(\mathbf{X}; \theta_{\text{att}}) + \text{FFN}(\text{Attn}(\mathbf{X}; \theta_{\text{att}}); \theta_{\text{ffn}}),
\end{equation}
with task-specific loss \( \mathcal{L}(\theta_{\text{att}}, \theta_{\text{ffn}}; \mathcal{M}) \).

As observed in Figure~\ref{pilot-attn} and~\ref{pilot-ffn},  when the base model is updated from \( \mathcal{M}_0 \) to \( \mathcal{M}_1 \), the attention sub-layers in both versions related to task-specific fine-tuning remain largely unchanged: \( \mathbf{W}_{\text{att}}^{(1)} \approx \mathbf{W}_{\text{att}}^{(0)} \). In contrast, the FFN sub-layers undergo significant parameter shifts. Therefore, we assume:

\begin{enumerate}
    \item \textit{Attention Stability}: $\|\mathbf{W}_{\text{att}}^{(1)} - \mathbf{W}_{\text{att}}^{(0)}\|_2 \leq \epsilon_{\text{att}} \ll 1$
    \item \textit{FFN Perturbation Boundedness}: $\|\mathbf{W}_{\text{ffn}}^{(1)} - \mathbf{W}_{\text{ffn}}^{(0)}\|_2 \leq \rho$
    \item \textit{Loss Smoothness}: $\mathcal{L}$ is \( L \)-Lipschitz in \( \mathbf{W}_{\text{att}}, \mathbf{W}_{\text{ffn}} \) and \( \beta \)-Lipschitz in $ \theta_{\text{ffn}}$
    \item \textit{Loss Decomposability}: The total loss can be decomposed into additive components from the attention and FFN sub-layers $\mathcal{L}(\theta; \mathcal{M}) = \mathcal{L}_{\text{att}}(\theta_{\text{att}}; \mathcal{M}) + \mathcal{L}_{\text{ffn}}(\theta_{\text{ffn}}; \mathcal{M})$
\end{enumerate}

Based on these assumptions, Trans-PEFT is designed to maintain the effectiveness of PEFT modules when transferred from \( \mathcal{M}_0 \) to $\mathcal{M}_1$ by preserving dependencies on stable attention sub-layers while adapting to FFN changes through controlled randomness. Below, we provide an upper bound on the loss discrepancy when applying the PEFT parameters fine-tuned on the $\mathcal{M}_0$ to the updated model $\mathcal{M}_1$ by Trans-PEFT.  
\newtheorem{theorem}{Theorem}
\begin{theorem} 
\textit{Let $\theta_{\text{att}}^{*(0)}$, $\theta_{\text{ffn}}^{*(0)}$ be the PEFT parameters fine-tuned on \( \mathcal{M}_0 \) with Trans-PEFT. For \( \mathcal{M}_1 \), the loss discrepancy satisfies:}  
\begin{equation}
\begin{aligned}
&\lvert\mathcal{L}(\theta^{*(0)}; \mathcal{M}_1) - \mathcal{L}(\theta^{*(0)}; \mathcal{M}_0)\rvert \leq \\
&\underbrace{L\rho}_{\text{FFN Shift}} + \underbrace{2\beta \|\theta_{\text{ffn}}^{*(0)} - \theta_{\text{ffn}}^{*(1)}\|}_{\text{Parameter Deviation}} + \underbrace{C \cdot (p_i + p_c)}_{\text{Regularization}},
\end{aligned}
\end{equation}
where \( \theta_{\text{ffn}}^{*(1)} \) is the hypothetical optimal PEFT parameter for \( \mathcal{M}_1 \). And the loss function satisfies smoothness and decomposability.
\end{theorem}
 
This bound reveals three key components affecting the transfer performance: First, the FFN shift term ($L\rho$) reflects the impact of base model updates; Second, the parameter deviation term measures the proximity between our transferred PEFT parameters and the optimal PEFT parameters for $\mathcal{M}_1$, which can be reduced as Trans-PEFT encourages PEFT modules to capture persistent patterns in attention; Third, the regularization term, which is controlled by Trans-PEFT, quantifies the performance penalty caused by its stochastic perturbations (e.g., masking and dropping). Specifically, Intra-layer Knowledge Masking reduces the model’s sensitivity to intra-layer dimension perturbations by \( p_i \), and Cross-layer Knowledge Dropping enhances parameter robustness against cross-layer variations by \( p_c \). Through adjustment of  \( p_i \) and \( p_c \), we balance learning on \( \mathcal{M}_0 \) and adaptability to \( \mathcal{M}_1 \). The complete proof is in Appendix~\ref{proof}.

\begin{table*}[t]
\centering
\scalebox{0.84}{
\begin{tabular}{@{}cccccccccc>{\columncolor{gray!20}}cc@{}}
\toprule
\textbf{Type} & \textbf{Method} & \textbf{ARC-c} & \textbf{SIQA} & \textbf{WinoG.} & \textbf{BoolQ} & \textbf{ARC-e} & \textbf{PIQA} & \textbf{OBQA} & \textbf{HellaS.} & \textbf{Avg.} \\ \midrule
\multirow{4}{*}{\textbf{LoRA}} 
& Fine-tune${_n}$ & 82.3 & 78.4 & 84.5 & 72.6 & 92.9 & 88.3 & 87.6 & 94.0 & 85.1 \\
 \cmidrule(l){2-11}
 & Fine-tune${_o}$ & 78.8 & 79.9 & 84.1 & 73.1 & 89.6 & 87.8 & 86.4 & 93.1 & 84.1 \\
 & Direct Trans.${_{o\rightarrow n}}$ & 83.8 & 72.3 & 78.9 & 18.1 & 92.3 & 86.7 & 85.0 & 35.4 & 69.1 \\
 & Trans-PEFT${_{o\rightarrow n}}$ & 84.2 & 80.1 & 85.2 & 63.3 & 92.9 & 87.3 & 88.2 & 93.1 & 84.3 \\
  \midrule
\multirow{4}{*}{\textbf{Adapter}}
  & Fine-tune${_n}$ & 87.1 & 80.2 & 84.6 & 72.3 & 95.3 & 90.3 & 91.8 & 95.0 & 87.1 \\
    \cmidrule(l){2-11}
 & Fine-tune${_o}$ & 86.5 & 80.6 & 85.3 & 75.3 & 93.7 & 90.1 & 88.4 & 95.0 & 86.9 \\
 & Direct Trans.${_{o\rightarrow n}}$ & 81.7 & 76.7 & 70.6 & 39.6 & 90.1 & 85.4 & 85.8 & 84.2 & 76.8 \\
 & Trans-PEFT${_{o\rightarrow n}}$ & 88.2 & 78.9 & 82.1 & 69.6 & 95.4 & 88.4 & 90.6 & 93.2 & 85.8 \\ 
 \midrule
\multirow{4}{*}{\textbf{DoRA}} 
 & Fine-tune${_n}$ & 87.9 & 79.4 & 86.0 & 74.5 & 96.0 & 90.2 & 93.2 & 95.0 & 87.8 \\
 \cmidrule(l){2-11}
 & Fine-tune${_o}$ & 84.9 & 81.6 & 86.3 & 74.9 & 93.8 & 90.0 & 90.2 & 95.3 & 87.1 \\
 & Direct Trans.${_{o\rightarrow n}}$ & 53.8 & 64.0 & 54.5 & 53.9 & 60.5 & 41.7 & 62.2 & 50.9 & 55.2 \\
 & Trans-PEFT${_{o\rightarrow n}}$ & 88.1 & 81.1 & 84.0 & 65.6 & 95.0 & 88.5 & 90.8 & 94.1 & 85.9 \\ \bottomrule
\end{tabular}
}
\caption{Results on commonsense reasoning tasks. We evaluate the transfer performance of various PEFT methods on Qwen models. ${{o\rightarrow n}}$ denotes directly applying PEFT modules fine-tuned on the old version to the new version without re-tuning.}
\label{cs}
\end{table*}
\begin{table*}[t]
\centering
\begin{tabular}{@{}cccccc>{\columncolor{gray!20}}cc@{}}
\toprule
\multicolumn{1}{l}{} & \multicolumn{1}{l}{} & \multicolumn{2}{c}{\textbf{Mathematical Reasoning}} & \multicolumn{2}{c}{\textbf{Code Generation}} & \multicolumn{1}{l}{} \\ \cmidrule(lr){3-4} \cmidrule(lr){5-6}
\textbf{Model} & \textbf{Method} & \textbf{GSM8K} & \textbf{MATH} & \textbf{HumanEval} & \textbf{MBPP} & \cellcolor{gray!20}\textbf{Average}

 \\ \midrule
\multirow{4}{*}{\begin{tabular}[c]{@{}c@{}}Qwen\\ $_{(2\rightarrow 2.5)}$\end{tabular}}
 & Fine-tune${_n}$ & 84.91 & 49.32 & 75.00 & 71.96 & 70.30 \\
 \cmidrule(l){2-7}
 & Fine-tune${_o}$ & 80.97 & 44.10 & 62.80 & 68.52 & 64.10 \\
 & Direct Trans.${_{o\rightarrow n}}$ & 40.03 & 32.08 & 75.61 & 69.84 & 54.39 \\
 & Trans-PEFT${_{o\rightarrow n}}$ & 84.61 & 47.66 & 79.27 & 74.07 & 71.40 \\ \midrule
\multirow{4}{*}{\begin{tabular}[c]{@{}c@{}}InternLM\\ $_{(2\rightarrow 2.5)}$\end{tabular}}
& Fine-tune${_n}$ & 79.61 & 39.08 & 63.41 & 64.81 & 61.73 \\
\cmidrule(l){2-7}
 & Fine-tune${_o}$ & 74.37 & 28.88 & 47.56 & 62.43 & 53.31 \\
 & Direct Trans.${_{o\rightarrow n}}$ & 39.50 & 25.34 & 59.76 & 62.70 & 46.83 \\
 & Trans-PEFT${_{o\rightarrow n}}$ & 77.56 & 38.86 & 64.02 & 63.49 & 60.98 \\ \bottomrule
\end{tabular}
\caption{Results on mathematical reasoning and code generation tasks. We evaluate the transfer performance on Qwen and InternLM base models with LoRA.}
\label{table-mathcode}
\end{table*}
\section{Experiments}
\subsection{Experimental Setup}
\noindent\textbf{Base Models.}
To thoroughly evaluate the transferability of our approach, we perform experiments involving seven base models from three different sources, all of which are updated through continual pre-training. These include representing general updates from Qwen2-7B to Qwen2.5-7B in the Qwen family, InternLM2-7B to InternLM2.5-7B in the InternLM family, and domain-specific updates from DeepSeek-7B \citep{deepseekai2024deepseekllmscalingopensource} to DeepSeek-Coder-7B-v1.5~\cite{guo2024deepseekcoderlargelanguagemodel} and DeepSeek-Math-7B \citep{shao2024deepseekmathpushinglimitsmathematical}. 

\noindent\textbf{Datasets.}
The datasets used in our experiments primarily cover three domains: commonsense reasoning, mathematical reasoning, and code generation. In the commonsense reasoning tasks, we follow \citet{hu-etal-2023-llm} by fine-tuning and evaluating the model on the CommonsenseQA dataset including BoolQ~\citep{clark-etal-2019-boolq}, PIQA~\citep{DBLP:journals/corr/abs-1911-11641}, SIQA~\citep{sap-etal-2019-social}, HellaSwag~\citep{zellers-etal-2019-hellaswag}, WinoGrande~\citep{DBLP:journals/corr/abs-1907-10641}, ARC-e, ARC-c~\citep{DBLP:journals/corr/abs-1803-05457}, and OBQA~\citep{mihaylov-etal-2018-suit} datasets. In the mathematical reasoning tasks, we fine-tune the model on a 100K subset of the MetaMathQA dataset \citep{yu2024metamath} and evaluate transfer performance on GSM8K \citep{DBLP:journals/corr/abs-2110-14168} and MATH \citep{DBLP:journals/corr/abs-2103-03874} datasets. In the code generation tasks, we fine-tune the model on the CodeFeedback105K dataset \citep{zheng2025opencodeinterpreterintegratingcodegeneration,meng2024pissa} and evaluate performance on HumanEval \citep{DBLP:journals/corr/abs-2107-03374} and MBPP \citep{DBLP:journals/corr/abs-2108-07732} datasets.

\noindent\textbf{Baselines and Settings.}
We evaluate our approach using three representative PEFT methods: LoRA \citep{DBLP:conf/iclr/HuSWALWWC22}, Adapter \citep{DBLP:conf/icml/HoulsbyGJMLGAG19}, and DoRA \citep{pmlr-v235-liu24bn}, which is a popular variant of LoRA. We compare our method with the following baselines: \textbf{Fine-tune${_o}$}, which denotes fine-tuning PEFT on the old version of the base model and evaluating it on the old version; \textbf{Fine-tune${_n}$}, which denotes fine-tuning PEFT on the new version and evaluating it on the new version, represents the ideal performance; and \textbf{Direct Transfer${_{o\rightarrow n}}$}, which fine-tuning PEFT on the old version and directly transferring to the new version without any re-tuning. Our proposed method, \textbf{Trans-PEFT${_{o\rightarrow n}}$}, applies two novel strategies to fine-tune PEFT on the old version, and then directly applies it to the new version without any re-tuning. More details are provided in Appendix~\ref{sec-imp}.
\subsection{Experiments on Different PEFT Types}
Table~\ref{cs} presents the experimental results of three different types of PEFT on the Qwen base models for commonsense reasoning tasks. As shown, the Trans-PEFT method consistently outperforms the Direct Transfer approach across all three PEFT types, achieving substantial improvements of 15.2\%, 9.0\%, and 30.7\% on LoRA, Adapter, and DoRA, respectively. Notably, the Direct Transfer method yields only 55.2\% average performance with DoRA and underperforms on multiple tasks, indicating its impracticality for real-world applications. In contrast, Trans-PEFT achieves performance comparable to that of fine-tuning on the new version across all PEFT types, without the need for re-tuning. This demonstrates its ability to reduce dependency on certain knowledge and encourages the PEFT module to focus on the invariants in the attention mechanism effectively.
\begin{figure}[t]
        \centering
	\subfigure[Mathematical Reasoning]{
	\includegraphics[width=0.48\linewidth]{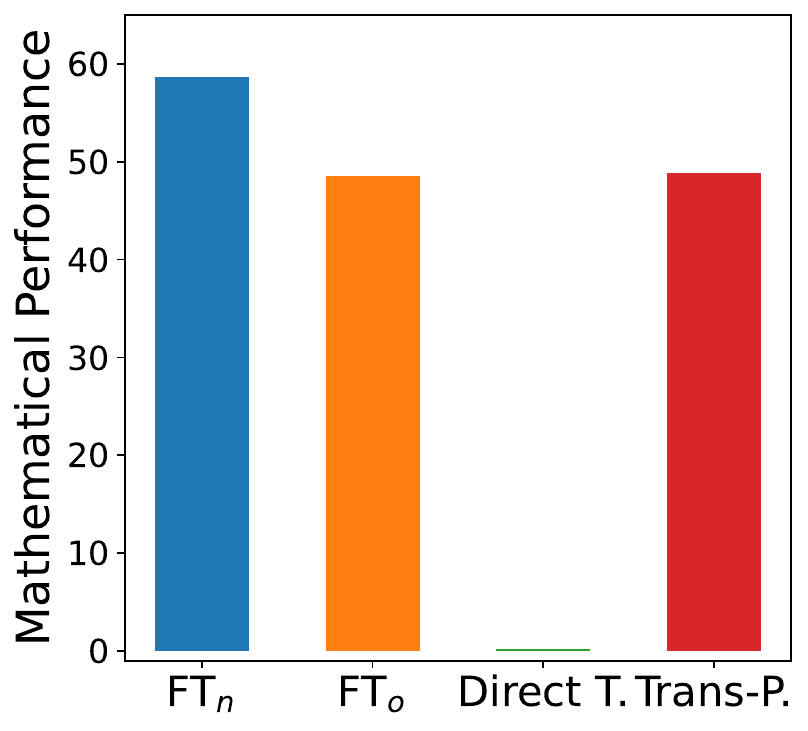}} 
        \subfigure[Code Generation]{
	\includegraphics[width=0.48\linewidth]{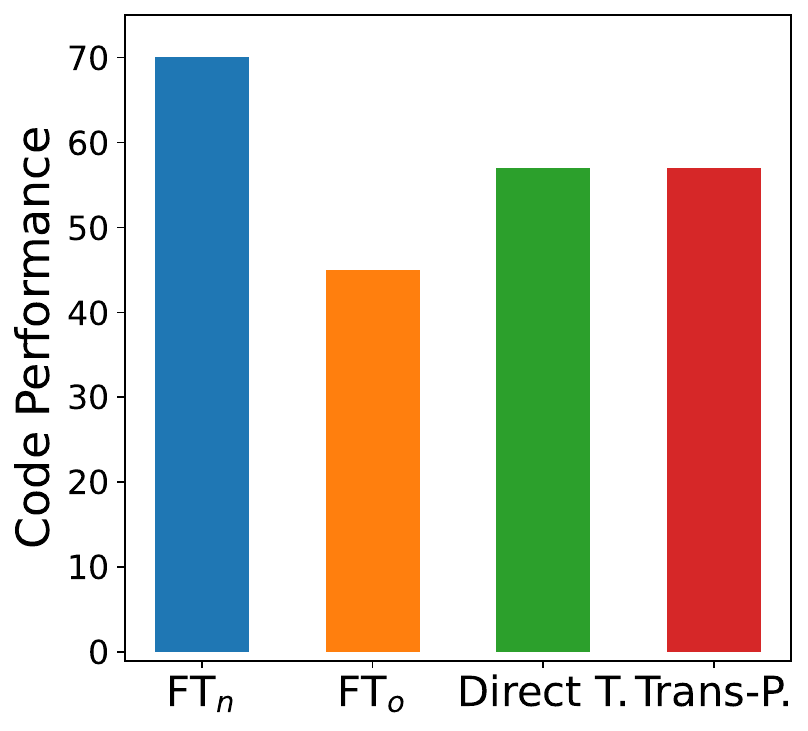}}
\caption{Results on base models with domain updates. We evaluate the transfer performance on DeepSeek base models with Adapter. For code tasks, we transfer from the base model to the code-updated version, while for math tasks, we transfer to the math-updated version.}
\label{domain results}
\end{figure}
\begin{figure}[t]
        \centering
	\subfigure[Effect of $p_c$]{
	\includegraphics[width=0.48\linewidth]{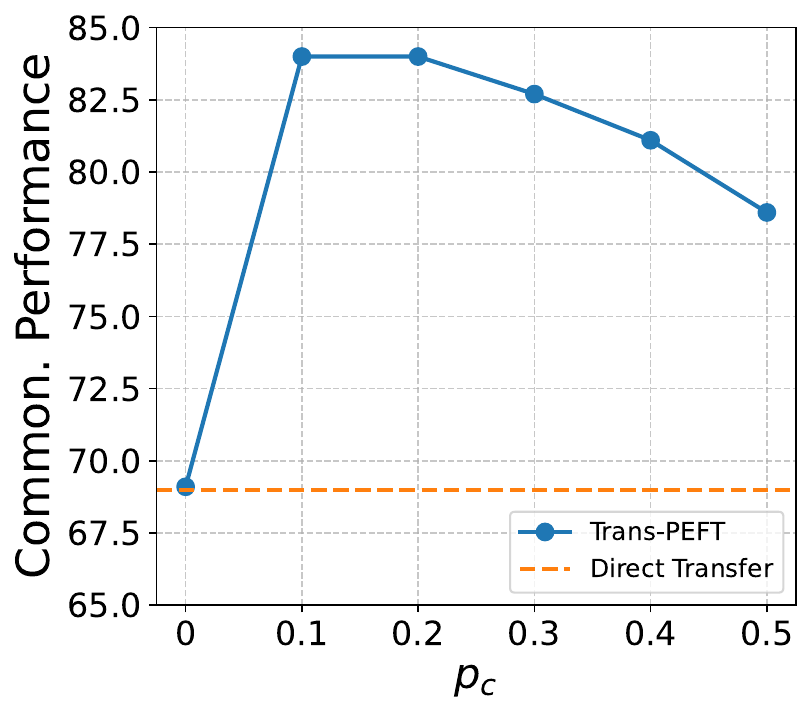}} 
        \subfigure[Effect of $p_i$]{
	\includegraphics[width=0.48\linewidth]{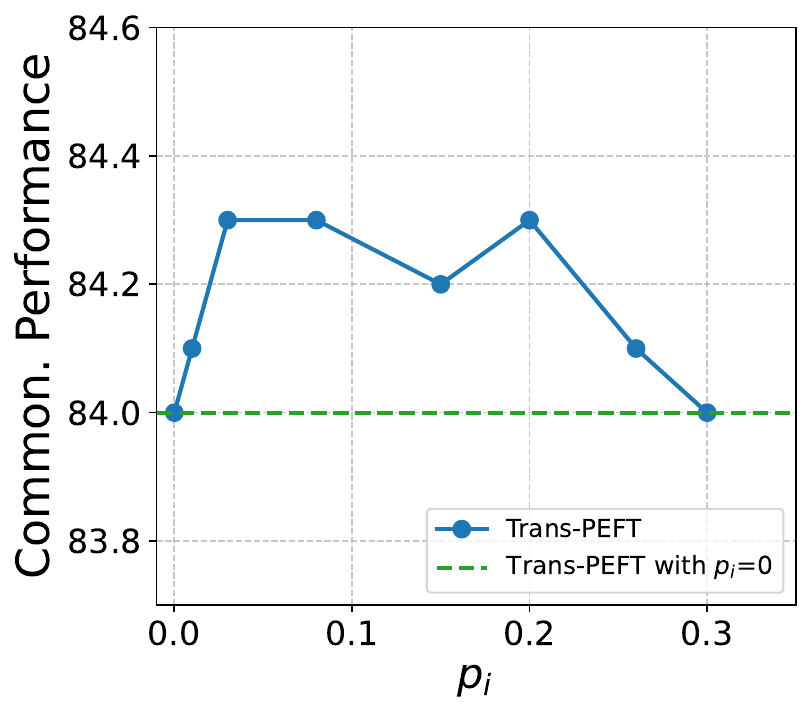}} 
\caption{Effects of $p_c$ and $p_i$ in our proposed strategies. The results are obtained on commonsense reasoning tasks using LoRA. In (a), we fix $p_i=0$ and vary $p_c$. In (b), we use the optimal setting $p_c=0.2$ and vary $p_i$.}
\label{ablation}
\end{figure}
\subsection{Experiments on Different Base Models}
Table~\ref{table-mathcode} presents results on different base models for mathematical reasoning and code generation tasks, which are generally considered more challenging. In commonsense reasoning experiments, the two base model versions show comparable fine-tuning performance, suggesting that base model updates do not significantly enhance commonsense reasoning capability. However, for mathematical and coding tasks, both Qwen and InternLM model families show significantly higher scores when fine-tuning on the new versions (i.e., Fine-tune${_n}$) compared to Fine-tune${_o}$, indicating that the base model updates specifically enhance capabilities in these domains. As shown, Trans-PEFT not only outperforms Fine-tune$_o$ in overall performance but also approaches the effectiveness of the re-tuning method Fine-tune$_n$, demonstrating its ability to leverage the performance gains from continual pre-training of the base model.  This success empirically validates our theoretical analysis, showing that Trans-PEFT can minimize the second term representing parameter deviation in fine-tuning processes. In contrast, the Direct Transfer approach not only fails to utilize the performance improvements from model updates but also underperforms compared to Fine-tune${_o}$. This suggests that the fine-tuned modules become ineffective after transfer.

Furthermore, we conduct experiments on the base model updated in the specific domain, as shown in Figure~\ref{domain results}. In mathematical and coding tasks, Trans-PEFT not only maintains the performance of the original model but also achieves improvements in coding tasks and slight enhancements in mathematical tasks. In comparison, Direct Transfer, while matching our performance in coding tasks, becomes nearly unusable after transferring to the more complex mathematical tasks. This demonstrates that our approach effectively captures task-specific patterns in attention mechanisms, enabling PEFT transfer across base model versions.

\subsection{Analysis}
\noindent\textbf{Effect of $p_c$ and $p_i$.}
We investigate the effect of two key parameters in Trans-PEFT: the layer drop probability \( p_c \) and dimension masking probability \( p_i \). These parameters control the regularization effects that reduce dependence on certain knowledge, thereby capturing invariant task-specific patterns. The parameter $p_c$ determines the probability of dropping each FFN layer during forward propagation. As illustrated in Figure~\ref{ablation}(a), when \( p_c \) is zero, neither of our proposed strategies is employed, resulting in suboptimal transfer performance. As \( p_c \) increases, transfer performance improves, peaking at \( p_c = 0.2 \). Further increases in $p_c$ begin to affect PEFT learning, leading to a decline in transfer performance.

The parameter $p_i$ controls the proportion of FFN dimensions masked during forward propagation.  As shown in Figure~\ref{ablation}(b), when $p_i = 0$, only the layer knowledge drop is applied. Moderate increases in $p_i$ further improve transfer performance compared to using layer knowledge drop alone, validating our dual-strategy approach. However, excessive $p_i$ values negatively affect PEFT learning, similar to the effects of high $p_c$ values.

\begin{figure}[t]
        \centering
	\subfigure[Adapter]{
	\includegraphics[width=0.31\linewidth]{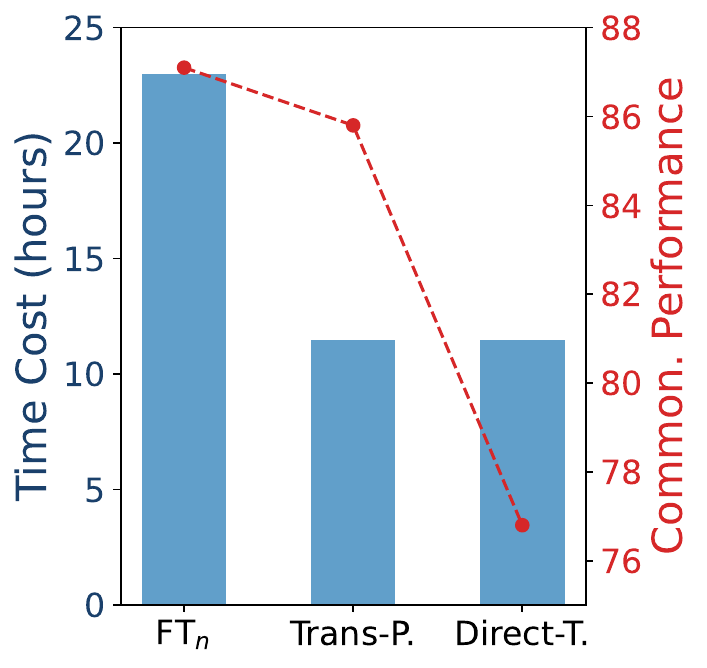}} 
        \subfigure[LoRA]{
	\includegraphics[width=0.31\linewidth]{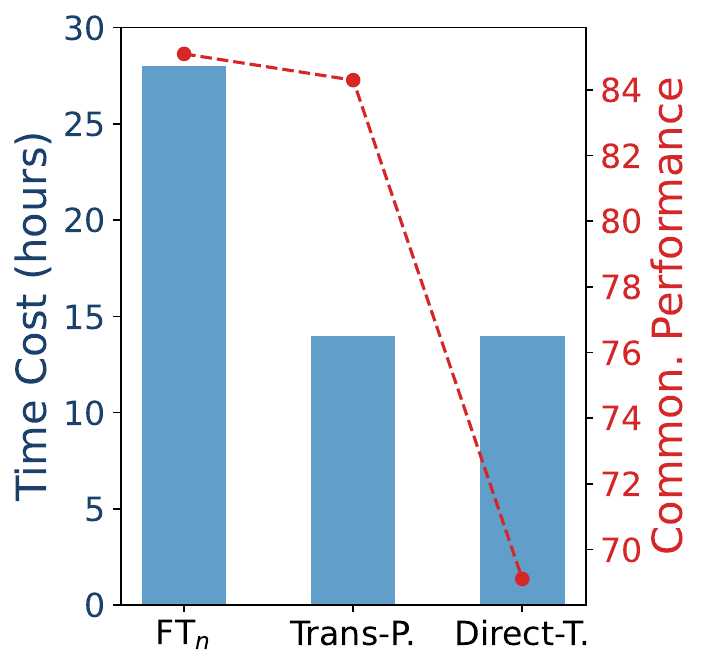}}
        \subfigure[DoRA]{
	\includegraphics[width=0.31\linewidth]{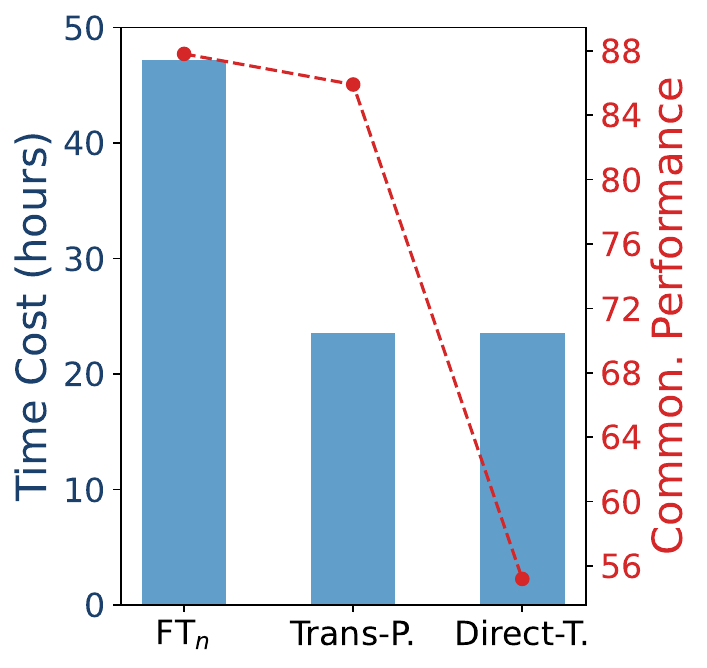}}
\caption{Fine-tuning time cost and performance of different methods across different types of PEFT on Qwen models. The fine-tuning time is tested on an A800 GPU, which includes the time overhead from fine-tuning on both the old and new base model versions.}
\label{eff}
\end{figure}

\noindent\textbf{Fine-tuning Efficiency.}
We conduct an analysis to assess the impact of Trans-PEFT on enhancing fine-tuning efficiency. As shown in Figure~\ref{eff}, fine-tuning different PEFT methods on the commonsense reasoning dataset typically takes 14 to 24 GPU hours on an A800 GPU. If re-tuning is conducted, including the initial fine-tuning time on the old version, the total time doubles to 28 to 48 hours (an additional 20 GPU hours on average). In practical applications, this efficiency gap becomes even more pronounced. For instance, when dealing with thousands of customized PEFT modules, the cumulative re-tuning time could escalate to approximately 20,000 GPU hours\footnote{It’s important to note that, unlike pre-training, batching and updating thousands of user-customized PEFT modules in parallel presents significant challenges. This is primarily due to the divergent objectives of different modules, necessitating multiple forward passes to update each module individually.}. Our Trans-PEFT approach offers a solution to this challenge by effectively eliminating the need for re-tuning while maintaining comparable performance levels. It requires only a single fine-tuning process on the old base model version, making it highly practical and efficient. This significant reduction in computational resources and time makes Trans-PEFT particularly valuable for large-scale applications and frequent model updates.

\section{Related Work}
\subsection{Parameter Efficient Fine-Tuning}
Parameter-efficient fine-tuning (PEFT) methods have several mainstream approaches to adding trainable modules:
Prompt-based methods \citep{lester-etal-2021-power,li-liang-2021-prefix,liu-etal-2022-p} fine-tune soft prompts added before the hidden states in model layers, but are less commonly used due to their relatively unstable performance and reduced input length \citep{han2024parameterefficient}. Adapter Tuning and its variants \citep{DBLP:conf/icml/HoulsbyGJMLGAG19,he2022towards} insert trainable modules between model layers, offering efficient parameter updates. Reparameterization methods, such as LoRA and its variants \citep{DBLP:conf/iclr/HuSWALWWC22,pmlr-v235-liu24bn,jiang2024morahighrankupdatingparameterefficient,gu2025beamlorabeamconstraintlowrankadaptation}, take a different approach by inserting low-rank modules parallel to the model's linear layers to approximate updates of the original weights. Both of these methods typically achieve performance comparable to full parameter fine-tuning. Building on these established approaches, we propose the Trans-PEFT method, which enables efficient transfer of fine-tuned PEFT modules across different versions of the base model while maintaining performance.
\subsection{Transfer Approaches on Updating Base Models}
The challenge of transferring PEFT modules across base models is first introduced by \citet{su-etal-2022-transferability}. Subsequent studies \citep{qin-etal-2023-recyclable,lester2022reducingretrainingrecyclingparameterefficient} discover that while the homologous model could reuse PEFT modules to some extent, there remain significant performance gaps compared to re-tuning, thereby limiting their practical utility. To bridge this gap, Trans-PEFT focuses on transferring PEFT modules between models updated via continual pre-training and introduces a novel transfer mechanism that maintains performance while reducing computational costs. From a data privacy perspective, \citet{wang2024textittransloradatafreetransferableparameter} proposes to eliminate the need for storing user data for re-tuning by using synthetic data generated from the old model version. While this approach preserves privacy, it still incurs substantial computational costs due to the need for re-tuning the PEFT modules. Trans-PEFT provides a more efficient solution by enhancing the fine-tuning process of the old version, which both reduces computational demands and eliminates the need for storing sensitive user data. Further discussion of related work is provided in Appendix~\ref{sec-extended-related-work}.

\section{Conclusion}
This paper introduces Trans-PEFT, a novel method that enables PEFT modules to work on updated base models. Our analysis reveals that base model updates primarily affect task-specific knowledge stored in FFN sub-layers while having minimal impact on task-specific patterns in attention mechanisms. Leveraging these insights, we develop both within-layer and cross-layer strategies to decrease PEFT modules’ dependence on FFN-stored knowledge, instead encouraging them to capture persistent patterns within the attention mechanisms. Extensive experiments show that Trans-PEFT successfully maintains performance across base model updates without requiring re-tuning, thereby significantly reducing the maintenance cost of PEFT.

\section*{Limitations}
Trans-PEFT demonstrates efficacy in transferring PEFT modules between continuously trained base models. However, our method faces constraints when applied to re-pretraining scenarios. Such scenarios often include architectural changes or large-scale dataset expansions, like updating from LLaMA2 \citep{touvron2023llama2openfoundation} to LLaMA3 \citep{grattafiori2024llama3herdmodels}. This limitation stems from the fundamental differences in pre-trained parameter spaces between base models with different random initializations \citep{imfeld2024transformer}.  Since PEFT operates within a specific parameter space~\cite{aghajanyan-etal-2021-intrinsic}, transferring parameters across completely distinct spaces remains infeasible. Exploring ways to adapt our method to these conditions represents a direction for future research.

\section*{Acknowledgments}
We would like to thank the anonymous reviewers for their valuable comments. We also thank Zhengxiao Liu from the Institute of Information Engineering for his help. This work is supported by the National Natural Science Foundation of China (No. 62472419, 62472420, 62376257).

\bibliography{anthology,custom}
\newpage
\appendix
\onecolumn
\section{Proof of Theorem 1}
\label{proof}
\subsection{Theorem Statement}
Let \( \mathcal{M}_0 \) and \( \mathcal{M}_1 \) denote two versions of a base transformer model, where \( \mathcal{M}_1 \) is updated from \( \mathcal{M}_0 \). We define \( \mathbf{W}^{(0)} = \{ \mathbf{W}_{\text{ffn}}^{(0)}, \mathbf{W}_{\text{att}}^{(0)} \} \) as the weights of \( \mathcal{M}_0 \), 
\( \theta^{(0)} = \{ \theta^{(0)}_{\text{ffn}}, \theta^{(0)}_{\text{att}} \} \) as the fine-tuned PEFT weights on \( \mathcal{M}_0 \); 
\( \mathbf{W}^{(1)} = \{ \mathbf{W}_{\text{ffn}}^{(1)}, \mathbf{W}_{\text{att}}^{(1)} \} \) as the weights of \( \mathcal{M}_1 \), 
\( \theta^{(1)} = \{ \theta_{\text{ffn}}^{(1)}, \theta_{\text{att}}^{(1)} \} \) as the fine-tuned PEFT weights on \( \mathcal{M}_1 \).

\noindent\textbf{Assumptions.}
\begin{enumerate}
    \item \textit{Attention Stability}: $\|\mathbf{W}_{\text{att}}^{(1)} - \mathbf{W}_{\text{att}}^{(0)}\|_2 \leq \epsilon_{\text{att}} \ll 1$
    \item \textit{FFN Perturbation Boundedness}: $\|\mathbf{W}_{\text{ffn}}^{(1)} - \mathbf{W}_{\text{ffn}}^{(0)}\|_2 \leq \rho$
    \item \textit{Loss Smoothness}: $\mathcal{L}$ is \( L \)-Lipschitz in \( \mathbf{W}_{\text{att}}, \mathbf{W}_{\text{ffn}} \) and \( \beta \)-Lipschitz in $ \theta_{\text{ffn}}$
    \item \textit{Loss Decomposability}: The total loss can be decomposed into additive components from the attention and FFN sub-layers $\mathcal{L}(\theta; \mathcal{M}) = \mathcal{L}_{\text{att}}(\theta_{\text{att}}; \mathcal{M}) + \mathcal{L}_{\text{ffn}}(\theta_{\text{ffn}}; \mathcal{M})$
\end{enumerate}
\noindent\textbf{Theorem 1.} \textit{Let $\theta^{*(0)}=\{\theta_{\text{att}}^{*(0)}, \theta_{\text{ffn}}^{*(0)}\} $ be the optimal PEFT parameters fine-tuned on \( \mathcal{M}_0 \) using Trans-PEFT. Under the assumptions of \textbf{attention stability} and \textbf{FFN perturbation boundedness}, the loss discrepancy on \( \mathcal{M}_1 \) satisfies:}
\begin{equation}
\begin{aligned}
    \lvert\mathcal{L}(\theta^{*(0)}; & \mathcal{M}_1) - \mathcal{L}(\theta^{*(0)}; \mathcal{M}_0)\rvert   \\
    &\leq L\rho + 2\beta \|\theta_{\text{ffn}}^{*(0)} - \theta_{\text{ffn}}^{*(1)}\| + C \cdot (p_i + p_c),
\end{aligned}
\end{equation}
\textit{where \( \theta_{\text{ffn}}^{*(1)} \) is the hypothetical optimal PEFT parameter for \( \mathcal{M}_1 \). And the loss function satisfies smoothness and decomposability.}
\subsection{Proof}
Since Trans-PEFT introduces a mask in the FFN module $\mathbf{\tilde{W}}_\text{ffn}^{(0)}=\mathbf{W}_{\text{ffn}}^{(0)} + \delta(m,z)$, we can therefore derive:

\begin{equation}
\begin{aligned}
    \lvert\mathcal{L}(\theta^{*(0)}&; \mathcal{M}_1)  - \mathcal{L}(\theta^{*(0)}; \mathcal{M}_0)\rvert \\
    &= \lvert\mathcal{L}(\theta^{*(0)}; [\mathbf{W}_\text{att}^{(1)},\mathbf{W}_\text{ffn}^{(1)}])- \mathcal{L}(\theta^{*(0)}; [\mathbf{W}_\text{att}^{(0)},\mathbf{\tilde{W}}_\text{ffn}^{(0)}])\rvert. \\
\end{aligned}
\end{equation}
The total loss can be separated into attention and FFN components (Assumption 4):
\begin{equation}
\begin{aligned}
    \lvert\mathcal{L}(\theta^{*(0)}& ; [\mathbf{W}_\text{att}^{(1)},\mathbf{W}_\text{ffn}^{(1)}]) - \mathcal{L}(\theta^{*(0)}; [\mathbf{W}_\text{att}^{(0)},\mathbf{\tilde{W}}_\text{ffn}^{(0)}])\rvert \\
    =& \lvert\mathcal{L}_{\text{att}}(\theta^{*(0)}_{\text{att}}; \mathbf{W}_\text{att}^{(1)}) + \mathcal{L}_{\text{ffn}}(\theta^{*(0)}_{\text{ffn}}; \mathbf{W}_\text{ffn}^{(1)}) - \mathcal{L}_{\text{att}}(\theta^{*(0)}_{\text{att}}; \mathbf{W}_\text{att}^{(0)}) - \mathcal{L}_{\text{ffn}}(\theta^{*(0)}_{\text{ffn}}; \mathbf{\tilde{W}}_\text{ffn}^{(0)})\rvert \\
    =& \lvert\mathcal{L}_{\text{att}}(\theta^{*(0)}_{\text{att}}; \mathbf{W}_\text{att}^{(1)}) - \mathcal{L}_{\text{att}}(\theta^{*(0)}_{\text{att}}; \mathbf{W}_\text{att}^{(0)}) + \mathcal{L}_{\text{ffn}}(\theta^{*(0)}_{\text{ffn}}; \mathbf{W}_\text{ffn}^{(1)})  - \mathcal{L}_{\text{ffn}}(\theta^{*(0)}_{\text{ffn}}; \mathbf{\tilde{W}}_\text{ffn}^{(0)})\rvert \\
    \leq& \lvert\mathcal{L}_{\text{att}}(\theta^{*(0)}_{\text{att}}; \mathbf{W}_\text{att}^{(1)}) - \mathcal{L}_{\text{att}}(\theta^{*(0)}_{\text{att}}; \mathbf{W}_\text{att}^{(0)})\rvert + \lvert\mathcal{L}_{\text{ffn}}(\theta^{*(0)}_{\text{ffn}}; \mathbf{W}_\text{ffn}^{(1)})  - \mathcal{L}_{\text{ffn}}(\theta^{*(0)}_{\text{ffn}}; \mathbf{\tilde{W}}_\text{ffn}^{(0)})\rvert. \\
\end{aligned}
\end{equation}
Since \( \mathbf{W}_{\text{att}}^{(1)} \approx \mathbf{W}_{\text{att}}^{(0)} \) (Assumption 1) and the $L_\text{att}$-Lipschitz continuity of \( \mathcal{L}_{\text{att}} \) (Assumption 3), the attention loss discrepancy is negligible:
\begin{equation}
\lvert\mathcal{L}_{\text{att}}(\theta^{*(0)}_{\text{att}}; \mathbf{W}_\text{att}^{(1)}) - \mathcal{L}_{\text{att}}(\theta^{*(0)}_{\text{att}}; \mathbf{W}_\text{att}^{(0)})\rvert \leq L_{\text{att}} \epsilon_{\text{att}} \approx 0.
\end{equation}
Thus, the total loss discrepancy is dominated by the FFN term:
\begin{equation}
\begin{aligned}
\lvert\mathcal{L}(\theta^{*(0)};& \mathcal{M}_1) - \mathcal{L}(\theta^{*(0)}; \mathcal{M}_0)\rvert \\
&\leq \lvert\mathcal{L}_{\text{ffn}}(\theta^{*(0)}_{\text{ffn}}; \mathbf{W}_\text{ffn}^{(1)})  - \mathcal{L}_{\text{ffn}}(\theta^{*(0)}_{\text{ffn}}; \mathbf{\tilde{W}}_\text{ffn}^{(0)})\rvert.
\end{aligned}
\end{equation}
Thus, we can derive the sub-terms of the loss:
\begin{equation}
\begin{aligned}
&\lvert\mathcal{L}_{\text{ffn}}(\theta_{\text{ffn}}^{*(0)}; \mathbf{W}_\text{ffn}^{(1)}) - \mathcal{L}_{\text{ffn}}(\theta_{\text{ffn}}^{*(0)}; \mathbf{\tilde{W}}_\text{ffn}^{(0)}) \rvert\\
    &\ \ = \lvert\mathcal{L}_{\text{ffn}}(\theta_{\text{ffn}}^{*(0)}; \mathbf{W}_\text{ffn}^{(1)})  - \mathcal{L}_{\text{ffn}}(\theta_{\text{ffn}}^{*(0)}; \mathbf{W}_\text{ffn}^{(0)}) + \mathcal{L}_{\text{ffn}}(\theta_{\text{ffn}}^{*(0)}; \mathbf{W}_\text{ffn}^{(0)}) - \mathcal{L}_{\text{ffn}}(\theta_{\text{ffn}}^{*(0)}; \mathbf{\tilde{W}}_\text{ffn}^{(0)})\rvert\\
   &\ \ \leq \lvert\mathcal{L}_{\text{ffn}}(\theta_{\text{ffn}}^{*(0)}; \mathbf{W}_\text{ffn}^{(1)})  - \mathcal{L}_{\text{ffn}}(\theta_{\text{ffn}}^{*(0)}; \mathbf{W}_\text{ffn}^{(0)})\rvert + \underbrace{\lvert\mathcal{L}_{\text{ffn}}(\theta_{\text{ffn}}^{*(0)}; \mathbf{W}_\text{ffn}^{(0)}) - \mathcal{L}_{\text{ffn}}(\theta_{\text{ffn}}^{*(0)}; \mathbf{\tilde{W}}_\text{ffn}^{(0)})\rvert}_{\text{sub-term }\mathcal{R}}\\
&\ \ \leq\underbrace{\lvert\mathcal{L}_{\text{ffn}}(\theta_{\text{ffn}}^{*(0)}; \mathbf{W}_\text{ffn}^{(1)}) - \mathcal{L}_{\text{ffn}}(\theta_{\text{ffn}}^{*(1)}; \mathbf{W}_\text{ffn}^{(1)})\rvert }_{\text{sub-term } \mathcal{A}_1}+\lvert\mathcal{L}_{\text{ffn}}(\theta_{\text{ffn}}^{*(1)}; \mathbf{W}_\text{ffn}^{(1)}) -\mathcal{L}_{\text{ffn}}(\theta_{\text{ffn}}^{*(0)}; \mathbf{W}_\text{ffn}^{(0)})\rvert +\mathcal{R}\\
&\ \ \leq\mathcal{A}_1 + \underbrace{\lvert\mathcal{L}_{\text{ffn}}(\theta_{\text{ffn}}^{*(1)}; \mathbf{W}_\text{ffn}^{(1)}) - \mathcal{L}_{\text{ffn}}(\theta_{\text{ffn}}^{*(1)}; \mathbf{W}_\text{ffn}^{(0)})\rvert}_{\text{sub-term }\mathcal{B}} + \underbrace{\lvert\mathcal{L}_{\text{ffn}}(\theta_{\text{ffn}}^{*(1)}; \mathbf{W}_\text{ffn}^{(0)})-\mathcal{L}_{\text{ffn}}(\theta_{\text{ffn}}^{*(0)}; \mathbf{W}_\text{ffn}^{(0)})\rvert}_{\text{sub-term }\mathcal{A}_2} +\mathcal{R}\\
&\ \ =\mathcal{A}_1 +\mathcal{B} + \mathcal{A}_2 +\mathcal{R}.
\end{aligned}
\end{equation}
Using the $L$-Lipschitz continuity of \( \mathcal{L}_{\text{ffn}} \) (Assumption 3) and the bound of FFN weights (Assumption 2), we bound the sub-term $\mathcal{B}$:
\begin{equation}
\begin{aligned}
\mathcal{B}=\lvert\mathcal{L}_{\text{ffn}}(\theta_{\text{ffn}}^{*(1)}; \mathbf{W}_\text{ffn}^{(1)})& - \mathcal{L}_{\text{ffn}}(\theta_{\text{ffn}}^{*(1)}; \mathbf{W}_\text{ffn}^{(0)})\rvert \\
&\leq L \cdot \|\mathbf{W}_{\text{ffn}}^{(1)} - \mathbf{W}_{\text{ffn}}^{(0)}\|_2 \leq L \rho.
\end{aligned}
\end{equation}
Using the \( \beta \)-Lipschitz in $ \theta_{\text{ffn}}$ (Assumption 3) we bound the sub-term $\mathcal{A}_1+\mathcal{A}_2$:
\begin{equation}
\begin{aligned}
\mathcal{A}_1+\mathcal{A}_2 =& \lvert\mathcal{L}_{\text{ffn}}(\theta_{\text{ffn}}^{*(0)}; \mathbf{W}_\text{ffn}^{(1)}) - \mathcal{L}_{\text{ffn}}(\theta_{\text{ffn}}^{*(1)}; \mathbf{W}_\text{ffn}^{(1)})\rvert +\lvert\mathcal{L}_{\text{ffn}}(\theta_{\text{ffn}}^{*(1)}; \mathbf{W}_\text{ffn}^{(0)})-\mathcal{L}_{\text{ffn}}(\theta_{\text{ffn}}^{*(0)}; \mathbf{W}_\text{ffn}^{(0)})\rvert \\
\leq & 2\beta \|\theta_{\text{ffn}}^{*(0)} - \theta_{\text{ffn}}^{*(1)}\|.
\end{aligned}
\end{equation}
On model $\mathcal{M}_0$, the fine-tuning objective of Trans-PEFT is to minimize the expected loss:
\begin{equation}
    \theta_{\text{ffn}}^{*(0)} = \arg\min_{\theta} \mathbb{E}_{m,z} \left[ \mathcal{L}_{\text{ffn}}(\theta; \mathbf{W}_{\text{ffn}}^{(0)} + \delta(m,z)) \right],
\end{equation}
where \( \delta(m,z) \) represents the implicit perturbation induced by masking \( m \) using probability \( p_i \) and dropping \( z \) using probability \( p_c \). Perform a second-order Taylor expansion of the loss function at the weight $ \mathbf{W}_{\text{ffn}}^{(0)}$:
\begin{equation}
   \mathcal{R} =  \lvert\mathcal{L}_{\text{ffn}}(\theta; \mathbf{W}_{\text{ffn}}^{(0)} + \delta)-\mathcal{L}_{\text{ffn}}(\theta; \mathbf{W}_{\text{ffn}}^{(0)})\rvert\approx \nabla_{\mathbf{W}} \mathcal{L}_{\text{ffn}} \cdot \delta + \frac{1}{2} \delta^\top \nabla_{\mathbf{W}}^2 \mathcal{L}_{\text{ffn}} \cdot \delta.
\end{equation}
After taking the expectation, because $\delta(m, z)$ is a zero-mean random variable, i.e., $\mathbb{E}_{m, z}[\delta(m, z)] = 0$, the first-order term vanishes. And the second-order term dominates the effect of the perturbation:
\begin{equation}
    \mathbb{E}[\mathcal{R}]\approx  \frac{1}{2} \mathbb{E}\left[ \delta^\top \nabla_{\mathbf{W}}^2 \mathcal{L}_{\text{ffn}} \cdot \delta \right].
\end{equation}
Assume the maximum eigenvalue of the Hessian matrix $\nabla_{\mathbf{W}}^2 \mathcal{L}_{\text{ffn}}$ is $\lambda_{\text{max}}$, then:
\begin{equation}
    \mathbb{E}\left[ \delta^\top \nabla_{\mathbf{W}}^2 \mathcal{L}_{\text{ffn}} \cdot \delta \right] \leq \lambda_{\text{max}} \cdot \mathbb{E}\left[ \|\delta\|_2^2 \right],
\end{equation}
where $\mathbb{E}\left[ \|\delta\|_2^2 \right]$ is introduced by Trans-PEFT via orthogonally designed strategies: intra-layer masking with probability $p_i$ and cross-layer dropping with probability $p_c$:
\begin{equation}
\mathbb{E}[\|\delta\|_2^2] = \mathbb{E}[\|\delta_i\|_2^2] + \mathbb{E}[\|\delta_c\|_2^2] \approx C_1 p_i + C_2 p_c,
\end{equation}
where $C_1$ and $C_2$ represent the sum of the squared norms of the masked weights by Trans-PEFT. Thus, we have the bound of $\mathcal{R}$:
\begin{equation}
\mathbb{E}\left[ \delta^\top \nabla_{\mathbf{W}}^2 \mathcal{L}_{\text{ffn}} \cdot \delta \right] \leq \lambda_{\text{max}} \cdot (C_1 p_i + C_2 p_c),
\end{equation}
\begin{equation}
\begin{aligned}
\mathbb{E}[\mathcal{R}] \leq \frac{1}{2}\lambda_{\text{max}} \cdot (C_1 p_i + C_2 p_c)\\ \leq C \cdot (p_i + p_c),
\end{aligned}
\end{equation}
where $C = \frac{1}{2} \lambda_{\text{max}} \cdot \max(C_1, C_2)$. Based on the above derivation, we obtain the three terms in the theorem.
\twocolumn
\section{Extended Related Work}
\label{sec-extended-related-work}
The challenge of cross-model transferability for parameter-efficient modules is first introduced by~\citet{su-etal-2022-transferability}, which observes that transferring prompt module parameters from the original model to a new model often leads to performance degradation or even total failure. The most straightforward solution to this problem is to restore performance through re-finetuning on the new model. MUSCLE~\cite{echterhoff2024musclemodelupdatestrategy} attempts to mitigate this by leveraging knowledge distillation to ensure that performance is maintained in the new model, while works such as Trans-LoRA~\cite{wang2024textittransloradatafreetransferableparameter} and Plug-n-Play~\cite{caccia2025trainingplugnplayknowledgemodules} further reduce or eliminate the need for additional downstream data for finetuning via distillation strategies. However, these methods ultimately still depend on re-finetuning on the new model and thus do not genuinely enable parameter transferability.

Existing studies aiming for module parameter transferability across models can be categorized into three directions: (1) cross-size transfer within the same model family, (2) transfer between entirely different model architectures, and (3) transfer between two models where one is updated from the other via continual training. The first direction, typified by works Offsite-Tuning~\cite{xiao2023offsitetuningtransferlearningmodel}, CRaSh~\cite{zhang-etal-2023-crash}, Plug-in~\cite{jin-etal-2023-parameter}, and LoRAM~\cite{DBLP:conf/iclr/ZhangW0SCYXGZ25}, tackles finetuning efficiency. Here, the modules are fine-tuned on a smaller, compressed version of the base model for efficiency and then transferred back to the original large model. These works mainly address issues of dimensionality mismatch rather than substantial parameter changes in the base model, making their approach orthogonal to ours. 

The second direction attempts to transfer prompt-based modules between two completely distinct models~\cite{lester2022reducingretrainingrecyclingparameterefficient,DBLP:conf/iclr/WuWM24,dong-etal-2025-contrans}, leveraging the small parameter size of prompt modules and alignment of overlapping vocabularies between the models. However, since the base models are independently pre-trained, there exist fundamental differences in their weight parameter spaces due to distinct initialization and training dynamics~\cite{imfeld2024transformer}. As PEFT methods operate within these model-specific parameter spaces~\cite{aghajanyan-etal-2021-intrinsic}, such approaches suffer from severely limited transferability and cannot be extended to mainstream PEFT methods like LoRA or to more complex tasks. 

The third direction addresses a common real-world scenario where models are incrementally updated via continual pre-training rather than re-pretraining from scratch. In this context, direct transfer represents the most straightforward solution. For instance, Chat Vector~\cite{huang-etal-2024-chat} and RE-Adapt~\cite{fleshman2024readaptreverseengineeredadaptation} both focus on direct transfer across fine-tuned base model versions, which involves limited continual training. For more substantial continual pre-training updates, Recycle Tuning~\cite{qin-etal-2023-recyclable} and PortLLM~\cite{DBLP:conf/iclr/ShahrozLYWNWC25} show that direct transfer between versions can outperform zero-shot learning on models updated through continual pre-training.
However, Recycle Tuning also finds that the performance of the direct transfer on newer versions is often much lower than that achieved by re-tuning, which is consistent with our findings. Our proposed method, Trans-PEFT, goes further by considering both the changes of knowledge during model updates in FFN sub-layers and the invariance of task-specific patterns in attention sub-layers, enabling the transferred parameter modules to maintain fine-tuned performance after the model is updated, and even achieve performance comparable to re-tuning.

From a methodological perspective, the Cross-layer Knowledge Dropping strategy in Trans-PEFT shares some similarities with LayerDrop~\cite{DBLP:conf/iclr/FanGJ20}, which focuses on inference efficiency; in contrast, our goal is to reduce the dependency of PEFT modules on specific sub-layers and to enhance cross-model transferability, marking a fundamental difference in motivation and outcome.
\section{Implementation Details}
\label{sec-imp}
\noindent\textbf{Observational Experiments.}
In the observational experiments, we fine-tune both the old and new versions of the model using LoRA with a rank of 64. To obtain the activations, we use a fixed subset for both fine-tuned models and obtain the activations via inference.

\noindent\textbf{Commonsense Reasoning Experiments.}
In commonsense reasoning experiments, we set the rank of all PEFT types to 32 and perform fine-tuning on the Commonsense170K dataset. The fine-tuning follows the settings of \citet{pmlr-v235-liu24bn}, with a batch size of 16 over three epochs. For all experiments, we set the random seed to 42. Finetune$_n$ and Finetune$_o$ denote typical fine-tuning and use on the corresponding versions. For Direct Transfer and Trans-PEFT, the PEFT modules are fine-tuned solely on the old version base model and then applied directly to the new version. For the Trans-PEFT parameters \(p_i\) and \(p_c\), given the large dataset, we select from \{0.1, 0.2, 0.3\}.

\noindent\textbf{Math Reasoning and Code Generation Experiments.} In mathematical reasoning and code generation experiments, we set the rank of all PEFT types to 64. Following the experimental setup of~\citet{meng2024pissa}’s setup, we fine-tune models on a 100K subset of MetaMathQA~\cite{yu2024metamath} and evaluate their 0-shot performance on GSM8K \citep{DBLP:journals/corr/abs-2110-14168} and MATH \citep{DBLP:journals/corr/abs-2103-03874} using the MetaMathQA codebase. For code generation tasks, we fine-tune models on the CodeFeedback105K dataset \citep{zheng2025opencodeinterpreterintegratingcodegeneration,meng2024pissa} and assess their performance on HumanEval \citep{DBLP:journals/corr/abs-2107-03374} and MBPP \citep{DBLP:journals/corr/abs-2108-07732} using the evalplus~\cite{evalplus}. For all experiments, we set the random seed to 42. As in commonsense experiments, we use a batch size of 16 and fine-tune over three epochs. For Trans-PEFT's parameter \(p_c\), we consider values from \{0.1, 0.2, 0.3\}, and \(p_i\) from \{0.01, 0.05, 0.1\}. All experiments are conducted on two A800 GPUs.

\section{Extended Observations}

\subsection{Observations on Other Tasks}
\begin{figure}[t]
        \centering
	\subfigure[InternLM2-7B]{
	\includegraphics[width=0.48\linewidth]{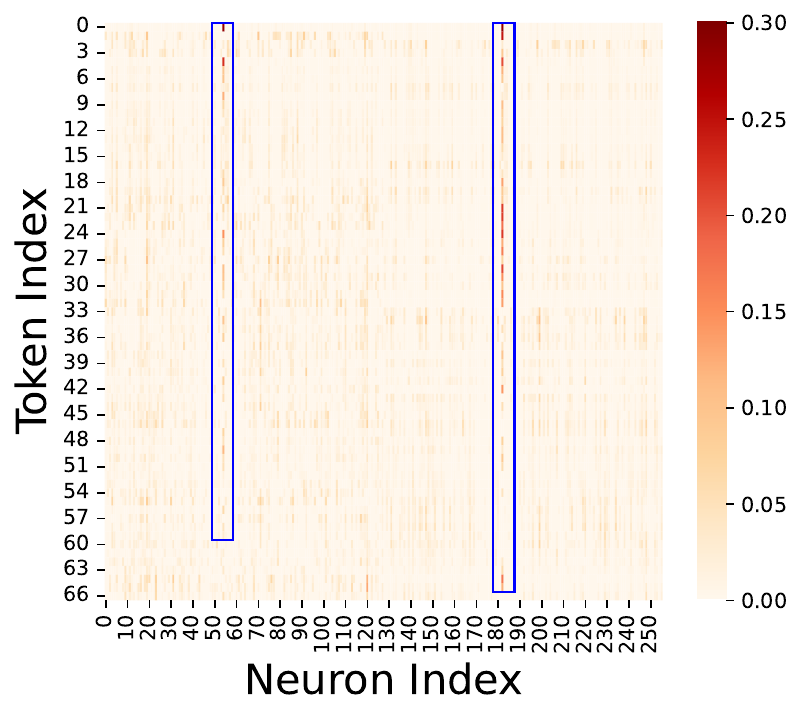}}
        \subfigure[InternLM2.5-7B]{
	\includegraphics[width=0.48\linewidth]{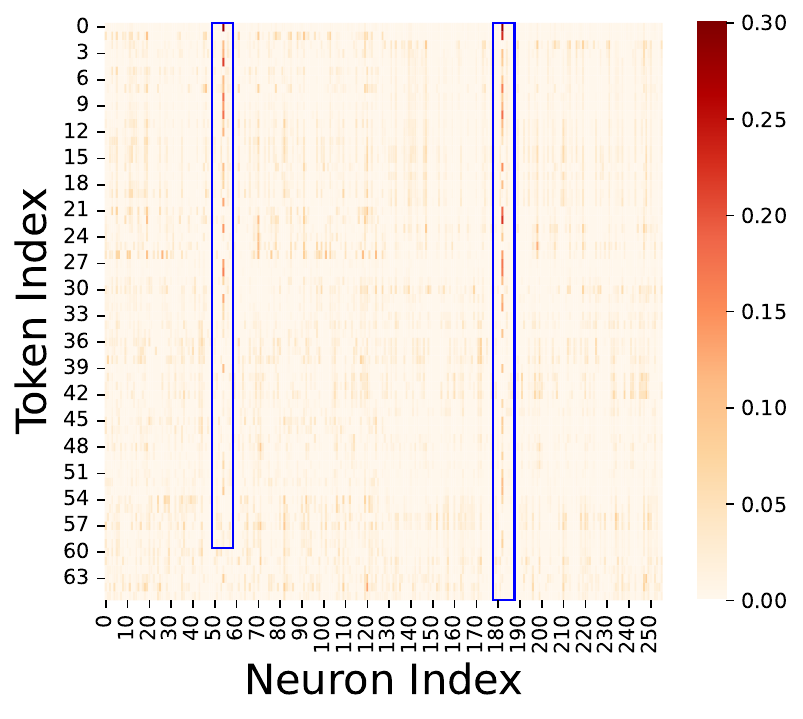}} 
\caption{Comparison of activation distributions within attention sub-layers across different fine-tuned model versions on the code task. The visualization shows the same dimensions from the 18th layer, with similar regularities observed across other layers.}
\label{app-pilot-attn}
\end{figure}
\begin{figure}[t]
        \centering
	\subfigure[InternLM2-7B]{
	\includegraphics[width=0.48\linewidth]{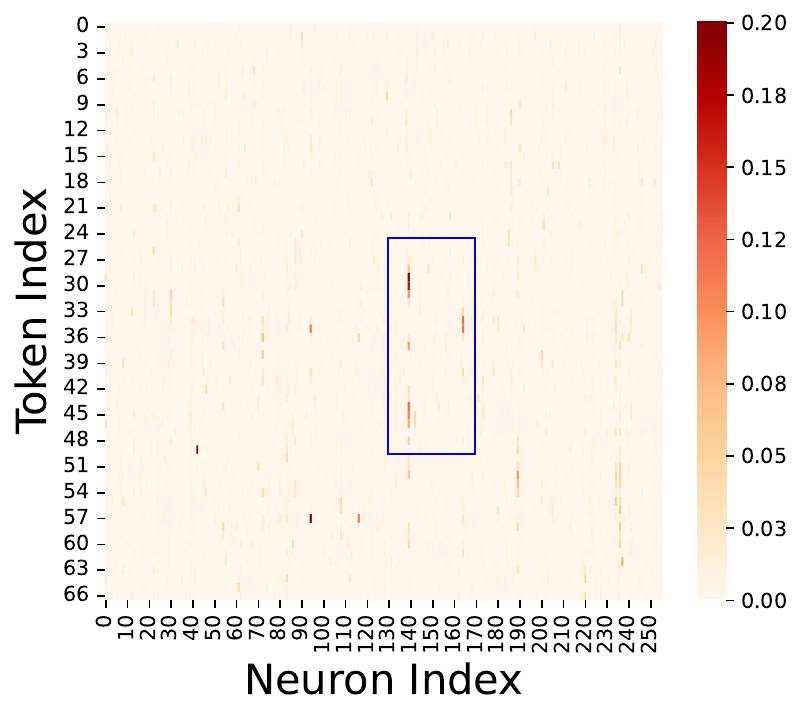}}
        \subfigure[InternLM2.5-7B]{
	\includegraphics[width=0.48\linewidth]{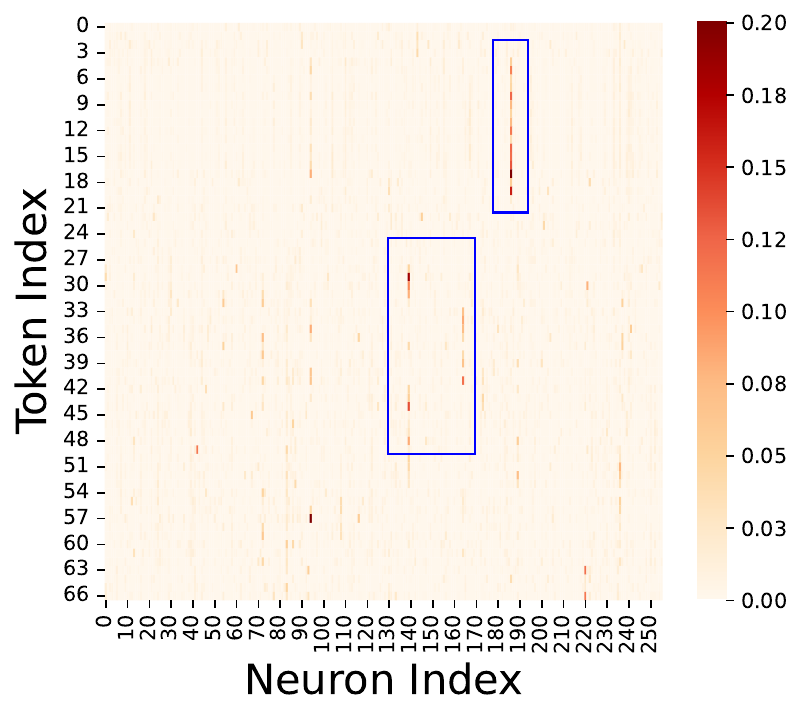}} 
\caption{Comparison of activation distributions within FFN sub-layers across different fine-tuned model versions on the code task. The visualization shows the same dimensions from the 18th layer, with similar regularities observed across other layers.}
\label{app-pilot-mlp}
\end{figure}

In Figures~\ref{app-pilot-attn} and~\ref{app-pilot-mlp}, we present the changes in activation distributions in the attention sub-layers and FFN sub-layers on the code generation task. As shown, the findings are consistent with those discussed in Section~\ref{sec-pilot}: task-specific pattern changes are less pronounced in the attention sub-layers, whereas knowledge storage changes are more significant in the FFN sub-layers.
\subsection{Further Discussion on the Impact of Model Updates}
\begin{figure}[t]
        \centering
	\includegraphics[width=0.8\linewidth]{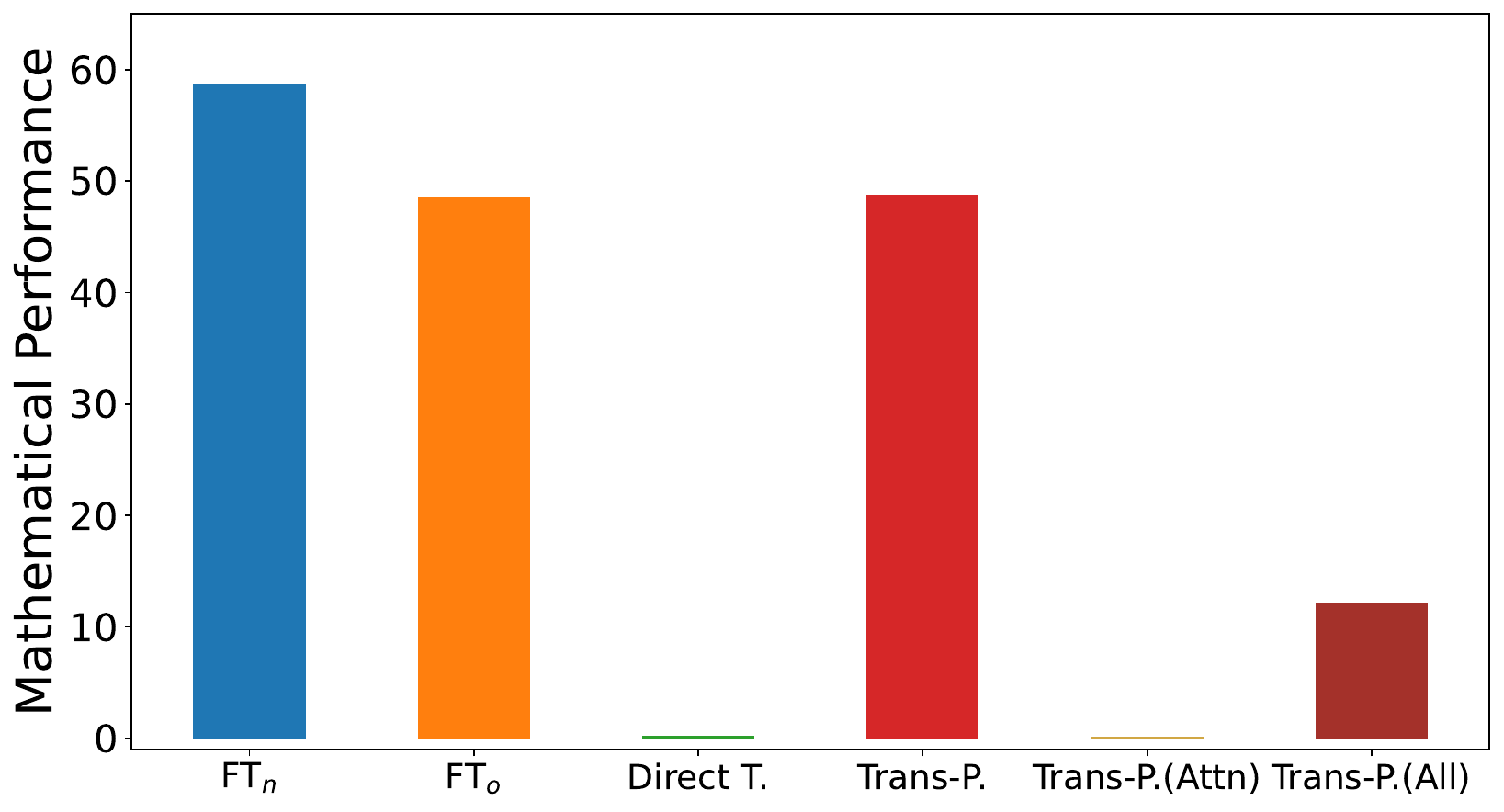}
\caption{Results on mathematical reasoning tasks with applying dropping on the attention sub-layer. We evaluate the transfer performance on DeepSeek base models with Adapter.}
\label{app-math}
\end{figure}
\begin{table*}[t]
\centering
\begin{tabular}{@{}cccccc@{}}
\toprule
 \textbf{} & \textbf{GSM8K} & \textbf{MATH} & \textbf{HumanEval} & \textbf{MBPP} & \textbf{Avg. Perf. Improv.} \\ \midrule
{Qwen$_{2\rightarrow2.5}$} & {p\textless{}0.01} & {p\textless{}0.01} & {p=0.03} & {p\textless{}0.01} & {17.01$_{\pm\text{0.84}}$} \\
{InternLM$_{2\rightarrow2.5}$} & {p\textless{}0.01} & {p\textless{}0.01} & {p=0.02} & {p=0.04} & {14.15$_{\pm\text{0.40}}$} \\ \bottomrule
\end{tabular}
\caption{Results of t-tests comparing Trans-PEFT with Direct Transfer on mathematical reasoning and code generation tasks.}
\label{table-t-test}
\end{table*}
\begin{table*}[t]
\centering
\begin{tabular}{@{}ccccc>{\columncolor{gray!20}}cc@{}}
\toprule
\multicolumn{1}{l}{} & \textbf{Method} & \textbf{seed 1} &  \textbf{seed 99} & \textbf{seed 1234} & \cellcolor{gray!20}\textbf{Average} \\ \midrule
 & Fine-tune${_n}$& 84.99 & 84.84 & 85.37 & 85.07 \\
 & Fine-tune${_o}$ & 80.67 & 81.43 & 81.05 & 81.05 \\
 & Direct Trans.${_{o\rightarrow n}}$ & 37.60 & 34.19 & 41.39 & 37.73 \\
\multirow{-4}{*}{GSM8K} & Trans-PEFT${_{o\rightarrow n}}$ & 84.76 & 85.14 & 84.91 & 84.94 \\ \midrule
 & Fine-tune${_n}$& 47.50 & 47.10 & 48.80 & 47.80 \\
 & Fine-tune${_o}$ & 44.04 & 43.86 & 42.16 & 43.35 \\
 & Direct Trans.${_{o\rightarrow n}}$ & 35.06 & 35.44 & 37.28 & 35.93 \\
\multirow{-4}{*}{MATH} & Trans-PEFT${_{o\rightarrow n}}$ & 47.76 & 45.74 & 48.78 & 47.43 \\ \bottomrule
\end{tabular}
\caption{Results on mathematical reasoning tasks with multiple seeds. We evaluate the transfer performance on Qwen base models with LoRA.}
\label{table-seed}
\end{table*}
To further validate our findings in Section~\ref{sec-pilot} that continual pre-training primarily affects task-specific knowledge stored in FFN while having minimal impact on task-specific patterns in attention mechanisms, we conduct additional experiments with Trans-PEFT applied to different sub-layers. Specifically, we apply Trans-PEFT to the attention sub-layer to examine the effects of reducing PEFT’s learning on these relatively stable task-specific patterns while maintaining its dependency on certain knowledge in FFN sub-layers. As shown in Figure~\ref{app-math}, applying Trans-PEFT solely to attention sub-layers yields similar poor transfer results as Direct Transfer, failing to adapt to the updated base model. Moreover, when Trans-PEFT is applied to both attention and FFN sub-layers, the transfer performance declines compared to applying it to the FFN sub-layer alone. This is because introducing randomness in attention sub-layers affects the capture of task-specific patterns. These quantitative results provide additional empirical support for our findings.

\section{Extended Experiments}
\subsection{Statistical Significance}
Table~\ref{table-t-test} presents the statistical significance results for the experiments in Table~\ref{table-mathcode}, where we fix the random seed to 42 and perform three runs with different data orders. As shown, the p-values for all tasks comparing Trans-PEFT and Direct Transfer are less than 0.05, indicating that Trans-PEFT significantly outperforms Direct Transfer. Moreover, the difference in average performance is also statistically significant. These results demonstrate that Trans-PEFT can effectively enhance transfer performance on different tasks.

\subsection{Stability Analysis of the Method}
In all of our experiments, we use a fixed seed (seed=42) for fine-tuning. In Table~\ref{table-seed}, we further assess the performance stability of Trans-PEFT across different seeds. The overall performance trends are consistent with those in Table~\ref{table-mathcode}, and Trans-PEFT achieves significantly better results than Direct Transfer. This demonstrates that the ability of Trans-PEFT to reduce dependence on task-specific knowledge in the FFN sub-layers is minimally affected by seed.

\end{document}